\documentclass{article}

\usepackage{PRIMEarxiv}

\usepackage{cite}
\usepackage[utf8]{inputenc} % allow utf-8 input
\usepackage[T1]{fontenc}    % use 8-bit T1 fonts
\usepackage{url}            % simple URL typesetting
\usepackage{booktabs}       % professional-quality tables
\usepackage{amsfonts}       % blackboard math symbols
\usepackage{amsmath}
\usepackage{nicefrac}       % compact symbols for 1/2, etc.
\usepackage{microtype}      % microtypography
\usepackage{lipsum}
\usepackage{fancyhdr}       % header
\usepackage{graphicx}
\usepackage{subcaption}
\usepackage[export]{adjustbox}
\usepackage{wrapfig}
\usepackage{algorithm}
\usepackage{algorithmic}
\usepackage{setspace}

\usepackage{tensor}  		%% /tensor{}{}

\newcommand{\mathtens}[1]{\tensor{\mathcal{#1}}{}}  	% using tensor package
\clearpage

\title{Decomposition of linear tensor transformations}

\author{
  Claudio Turchetti \\
  DII - Department of Information Engineering \\
  Università Politecnica delle Marche\\
  via Brecce Bianche,12, 60131 Ancona, Italy\\
  \texttt{c.turchetti@univpm.it} \\
}

\begin{document}

\setlength{\parindent}{20pt}

\large

\maketitle

\begin{abstract}
One of the main issues in computing a tensor decomposition is how to choose the number of rank-one components, since there is no finite algorithms for determining the rank of a tensor.  A commonly used approach for this purpose is to find a low-dimensional subspace by solving an optimization problem and assuming the number of components is fixed. However, even though this algorithm is efficient and easy to implement, it often converges to poor local minima and suffers from outliers and noise. The aim of this paper is to develop a mathematical framework for exact tensor decomposition that is able to represent a tensor as the sum of a finite number of low-rank tensors. In the paper three different problems will be carried out to derive: i) the decomposition of a non-negative self-adjoint tensor operator; ii) the decomposition of a linear tensor transformation; iii) the decomposition of a generic tensor.
\end{abstract}

% keywords can be removed

% ###### SECTION 1 ##########
%#########   INTRODUCTION  
%###############################

\section{Introduction}

\indent Tensors are the higher order generalization of vectors and matrices, and can consequently be treated as arrays indexed by multiple indices \cite{smilde2005multi},\cite{de1998matrix}, 
\cite{acar2008unsupervised}.
Thanks to their ability to represent a wide range of real-world data, tensors have expanded quickly  from psychometrics \cite{tucker1966some}, \cite{carroll1970analysis} and chemometrics
\cite{appellof1981strategies}, \cite{smilde2005multi} to image analysis  \cite{sofuoglu2021multi}, \cite{qin2022low}, \cite{wang2021multi}, \cite{sun2022tensor},  \cite{long2021bayesian}, \cite{zhang2020robust}, \cite{tian2022low}, big data representation \cite{kaur2018tensor}, machine learning, \cite{ji2019survey},
 \cite{signoretto2014learning}, \cite{hou2017fast}, \cite{lebedev2014speeding}, \cite{kim2015compression}, \cite{lin2018holistic}, sensor array processing \cite{sidiropoulos2000parallel}, and much more \cite{sankaranarayanan2015tensor},
 \cite{omberg2007tensor}, \cite{zhou2013tensor},
 \cite{sankaranarayanan2015tensor},\cite{liu2012tensor}, \cite{han2022rank}.
As the order of tensor increases, the number of entries in the array increases exponentially thus involving prohibitively computational and storage costs. To prevent these limitations, techniques for dimensionality reduction are required to make possible the application of tensors to real data. A widely adopted tecnique for this purpose is to represent high-dimensional tensors as linear combination of low-dimensional tensors. If $\mathtens{A}$ is an order-$d$ tensor, to reach this goal is essential to find low-order tensors $\mathtens{U}_i$ so that
\begin{equation}\label{sum}
\mathtens{A}=\sum_{{i}=1}^{{r}}\sigma_i\mathtens{U}_i,
\end{equation}
where $\sigma_i$ are positive scalars. In this way a low-dimensional representation can be derived by simply retaining the principal components of such a decomposition, that is the ones corresponding to the highest values of $\sigma_i$.
In this context several very powerful tensor decomposition approaches have been developed in the past \cite{kolda2009tensor}, \cite{cichocki2015tensor}, \cite{sidiropoulos2017tensor}, \cite{chen2021introduction}, in which tensors $\mathtens{U}_i$ in the summation (\ref{sum}) are rank-one tensors. Among these, Tucker decomposition \cite{tucker1963implications} and CANDECOMP/PARAFAC (CP) decomposition \cite{harshman1970foundations}, \cite{carroll1970analysis} are the most popular and fundamental models. Tucker model decomposes a tensor into a core tensor multiplied by a factor matrix along each mode, while CP model factorizes a tensor into a weighted sum of rank-1 tensors. 
Following these two seminal approaches, several extensions for low rank tensor decomposition have been proposed in the past few decades. Some of the most relevant developments in this field are: multilinear principal component analysis (MPCA)
\cite{inoue2016generalized}, \cite{liu2010generalized}, \cite{lu2008mpca}, \cite{mavzgut2014dimensionality}, \cite{panagakis2009non}, \cite{sun2008incremental}, \cite{xu2008reconstruction}, \cite{yang2004two}, \cite{ye2004generalized}, \cite{ye2004gpca}, 
tensor rank-one decomposition (TROD)
\cite{bro1997parafac}, \cite{faber2003recent}, \cite{harshman1970foundations}, \cite{kruskal1977three}, \cite{shashua2001linear} , 
hierarchical Tucker PCA (HT-PCA)
\cite{grasedyck2010hierarchical},  \cite{grasedyck2013literature}, \cite{hackbusch2009new}
and tensor-train PCA (TT-PCA)
\cite{bengua2017matrix}, \cite{chaghazardi2017sample}.\\
\indent
One of the main issues in computing a tensor decomposition is how to choose the number of rank-one components, since there is no finite algorithms for determining the rank of a tensor \cite{kruskal1977three}, \cite{haastad1989tensor}, defined as the smallest number of rank-one tensors that generate $\mathtens{A}$ as their sum. As a consequence, all the above methods for tensor decomposition try to find a low-dimensional subspace by solving an optimization problem and assuming the number of components is fixed. A commonly used approach for this purpose is the alternating least square (ALS) method \cite{zare2018extension}, which assumes the solution of all but one mode is known and then estimating the unknown parameter set of the remaining mode. However, even though this algorithm is efficient and easy to implement, it often converges to poor local minima and suffers from outliers and noise \cite{cheng2016probabilistic}.\\
\indent
The aim of this paper is to develop a mathematical framework for exact tensor decomposition that is able to represent a tensor as the sum of a finite number of low-rank tensors. The core of the proposed approach is the derivation of a decomposition for \textit{non-negative self-adjoint tensor operators}. In particular it will be proven that a correspondence exists between the spectral decomposition of a symmetric matrix, a result that is the basis of PCA and SVD development in matrix analysis, and the decomposition of a self-adjoint tensor operator. In the paper three different problems will be carried out to derive: i) the decomposition of a non-negative self-adjoint tensor operator; ii) the decomposition of a linear tensor transformation; iii) the decomposition of a generic tensor. In particular, with reference to the first issue, the properties of a self-adjoint tensor operator will be studied and the equivalence between eigenvalue equation for this operator and standard matrix eigenvalue equation will be proven.\\ 
\indent
The paper is organized as follows. Section 2 deals with the finite linear space of tensors and some fundamental concepts on linear tensor transformations. In Section 3 the mathematical framework for exact tensor decomposition is developed.
Section 4 reports some numerical results to validate the mathematical framework previously presented.

% ########SECTION 2 ############
%%%%%%%%%%%%%%%%%%%%%%%%%%%%%%%%%%%%%%%%%%%%%%%%%%%%%%%%%%%%%%%%%%%%%%%%%%%%%%%%%%
% The linear space of tensors
%%%%%%%%%%%%%%%%%%%%%%%%%%%%%%%%%%%%%%%%%%%%%%%%%%%%%%%%%%%%%%%%%%%%%%%%%%%%%%%%%%
\setlength{\parindent}{0pt}
\section{The linear space of tensors}
\label{sec:preliminaries}

Let us refer to the set $\mathtens{L}_d$ of \textit{real order-$d$ tensors} 
$\mathtens{X} \in \mathbb{R}^{I_1 \times I_2 \times \ldots\times{I_d}}$ whose  generic element will be denoted as

%%% eq.2.1
\begin{equation}\label{eq2.1}
\mathtens{X}_{\footnotesize\textbf{i}}=\mathtens{X}_{i_1, \ldots, i_d}, \;\; \textbf{i} = {(i_1, \ldots, i_d)}  \quad .
\end{equation}
%%%%%%%%%

The vector $\textbf{i}$, in bold font, is a \textit{subscript vector} where the index $i_k$ in the $k$th mode ranges in the interval
$1\leq{i_k}\leq{I_k},\hspace{0.2cm}  k=1:d$. As $\mathtens{L}_d$ is closed under addition and multiplication by scalar, this set forms a linear space, called the \textit{tensor space}. In the following we will refer to real tensors alone, thus for simplicity the term real will be omitted being always implied.

%%%%%%%%%%%%%%%%%%%%%%%%%%%%%%%%%%
% Inner product
%%%%%%%%%%%%%%%%%%%%%%%%%%%%%%%%%%
\paragraph{Inner product}  The inner product of two tensors of the same size $\mathtens{X}_{\footnotesize\textbf{i}}, 
\mathtens{Y}_{\footnotesize\textbf{i}} 
\in \mathbb{R}^{I_1 \times I_2 \times \ldots I_d}$ is a real scalar defined as

%%%%%%% eq.2.2
\begin{equation}\label{eq2.2}
\left\langle \mathtens{X}, \mathtens{Y} \right\rangle = \mathtens{X}_{\footnotesize\hspace{0.02cm}\textbf{i}}\mathtens{Y}_{\footnotesize\hspace{0.02cm}\textbf{i}}.
\end{equation}

Here the Einstein summation convention, that interprets repeated subscript as summation over that index, has been used. Following this convention, (\ref{eq2.2}) is equivalent to 

%%%%%%%%%%%% eq.2.3
\begin{equation}\label{eq2.3}
\left\langle \mathtens{X}, \mathtens{Y} \right\rangle = 
\sum_{{i_1}=1}^{{I_1}}...\sum_{{i_d}=1}^{{I_d}}
\mathtens{X}_{i_1, \ldots, i_d}
\mathtens{Y}_{i_1, \ldots, i_d}
\end{equation}

In the paper we will extensively use the Einstein convention to simplify mathematical notation.
From this definition $\mathtens{X},\mathtens{Y}$ are said to be orthogonal if $\left\langle \mathtens{X}, \mathtens{Y} \right\rangle=0$, and it follows that the norm of a tensor is given by 

%%%%% eq.2.4
\begin{equation}\label{eq2.4}
\|{\mathtens{X}}\| = \sqrt{\left\langle \mathtens{X}, \mathtens{X} \right\rangle}
\end{equation}

%%%%%%%%%%%%%%%%%%%%%%%%%%%%%%%%%%%%
%  Subspace
%%%%%%%%%%%%%%%%%%%%%%%%%%%%%%%%%%%%
\paragraph{Subspace}  A \textit{subspace} $S$ of $\mathbb{R}^{I_1 \times I_2 \times \ldots\times{I_d}}$ is a subset that is also a tensor space. Given a collection of tensors
$\mathtens{U}_1,\dots\mathtens{U}_n
\in\mathbb{R}^{I_1 \times I_2 \times \ldots\times{I_d}}$, the set of all linear combinations of these vectors is a subspace referred to as the 
$span\{\mathtens{U}_1,\dots\mathtens{U}_n\}$:

\begin{equation}\label{eq2.5}
span\{\mathtens{U}_1,\dots\mathtens{U}_n\}=
\{\alpha_i\mathtens{U}_i:\hspace{0.2cm}
\alpha_i\in\mathbb{R},\hspace{0.2cm}
i=1:n\}.
\end{equation}

%%%%%%%%%%%%%%%%%%%%%%%%%%%%%%%%%%%%%%%%%%%
% contraction
%%%%%%%%%%%%%%%%%%%%%%%%%%%%%%%%%%%%%%%%%%
\paragraph{Contraction} Given two tensors

%%%% eq.2.6
\begin{equation}\label{eq2.6}
\mathtens{X}_{\footnotesize\hspace{0.02cm}\textbf{i}}
\in \mathbb{R}^{I_1 \times I_2 \times \ldots\times{I_d}}, \hspace{0.5cm} 
 \mathtens{Y}_{\footnotesize\hspace{0.02cm}\textbf{j}}
 \in \mathbb{R}^{J_1 \times J_2 \times \ldots\times{J_e}}
\end{equation}
%%%%%%%%

and assuming a common vector index $\textbf{k}$ exists such that  ${\textbf{i}}$, ${\textbf{j}}$ can be partitioned as 

%%%% eq.2.7
\begin{equation}\label{eq2.7}
    \textbf{i}=(\textbf{l},\textbf{k},\textbf{m}), \hspace{0.5cm}\textbf{j}=(\textbf{p},\textbf{k},\textbf{q})
\end{equation}
%%%%%%%%

the tensor product of $\mathtens{X}$ and $\mathtens{Y}$ along the multi-index $\textbf{k}$ combines the two tensors to give a third tensor $\mathtens{Z}$, whose generic element is given by

%%%% eq.2.8
\begin{equation}\label{eq2.8}
\mathtens{Z}_{\footnotesize\hspace{0.02cm}\textbf{l},\textbf{m},\textbf{p},\textbf{q}}=
\mathtens{X}_{\footnotesize\hspace{0.02cm}\textbf{l},\textbf{k},\textbf{m}}
\mathtens{Y}_{\footnotesize\hspace{0.02cm}\textbf{p},\textbf{k},\textbf{q}}
\end{equation}

that implies summation over the vector index $\textbf{k}$. We refer to this operation  as \textit{contraction} of index \textbf{k}.
\newline

%%%%%%%%%%%%%%%%%%%%%%%%%%%%%%%%%%
% Outer product
%%%%%%%%%%%%%%%%%%%%%%%%%%%%%%%%%%
\paragraph{Outer product} The outer product of tensor $\mathtens{X}_{\footnotesize\textbf{i}} \in \mathbb{R}^{I_1 \times I_2 \times \ldots \times I_p}$ with
tensor $\mathtens{Y}_{\footnotesize\textbf{j}} \in \mathbb{R}^{J_1 \times J_2 \times \ldots \times J_q}$ is the order-$(p+q)$ tensor $\mathtens{Z}$
defined as

%%%%% eq.2.9
\begin{equation}\label{eq2.9}
\mathtens{Z} = \mathtens{X} \circ \mathtens{Y},
\end{equation}

where the generic element of $\mathtens{Z}$ is 

%%%%% eq.2.10
\begin{equation}\label{eq2.10}
\mathtens{Z}_{\footnotesize\textbf{i},\textbf{j}} = 
\mathtens{X}_{\footnotesize\textbf{i}} \mathtens{Y}_{\footnotesize\textbf{j}},
\hspace{0.3cm} \textbf{i} = {(i_1, \ldots, i_p)},\hspace{0.2cm}\textbf{j}={(j_1, \ldots, j_q)}
\end{equation}

In particular for two vectors $x,y\in{R^{I_1}}$ the outer product is the matrix ${Z} = x \circ y$ whose generic element is

%%%%% eq.2.11
\begin{equation}\label{eq2.11}
z_{ij} = x_{i} \, y_{j} \quad .
\end{equation}

%%%%%%%%%%%%%%%%%%%%%%%
% Vector-to-linear index transformation
%%%%%%%%%%%%%%%%%%%%%%%
\paragraph{Vector-to-linear index transformation} With reference to an order-$(p+q)$ tensor

%%%%%%%% eq.2.12
\begin{equation}\label{eq2.12}
{\mathtens{X}}_{\footnotesize\textbf{j},\textbf{i}},\hspace{0.2cm}\textbf{j}
={(j_1, \ldots, j_p)},\hspace{0.2cm} \textbf{i} = {(i_1, \ldots, i_q)}
\end{equation}

it is worth to notice that the indices $i_1, \ldots, i_q$ can be arranged in $M=I_1I_2\dots I_q$ different ways (the number of permutations with repetition), so that a correspondence $\alpha$ 

%%%%%%%% eq.2.14
\begin{equation}\label{eq2.14}
m=\alpha(\textbf{i}),\hspace{0.2cm}1\leq{m}\leq{M},
    \hspace{0.2cm}1\leq{i_k}\leq{I_k},
    \hspace{0.2cm}k=1:q
\end{equation}
between $m$ and $\textbf{i}$ holds.

The inverse transformation exists, since the correspondence is one-to-one, and is denoted by 

%%%%%%%% eq.2.15
\begin{equation}\label{eq2.15}
    \textbf{i}=\alpha^{-1}(m),\hspace{0.2cm}1\leq{m}\leq{M},\hspace{0.5cm}1\leq{i_k}\leq{I_k},
    \hspace{0.2cm}k=1:q
\end{equation}

By defining the $(M\times q)$ matrix $T$ whose $m$th row represents the corresponding vector index $\textbf{i}(m)$, then the inverse transformation  (\ref{eq2.15}) is formally given by

%%%% eq.2.17
\begin{equation}\label{eq2.17}
    \textbf{i}=T(m,:),\hspace{0.2cm}1\leq{m}\leq M
\end{equation}

As a result this operation, transforming the vector index $\textbf{i}$ to the linear index $m$, turns the order-$(p+q)$ tensor to the order-$(p+1)$ tensor 

%%%%%%%%% eq.2.13
\begin{equation}\label{eq2.13}
{\mathtens{X}}_{\footnotesize\textbf{j},\large{m}},
\hspace{0.2cm}\textbf{j}={(j_1, \ldots, j_p)},
\hspace{0.2cm} m = 1:M.
\end{equation}

The vector-to-linear-index transformation can be applied to more than one index. As an example, by transforming the two vector indices of $\mathtens{X}_{\footnotesize\textbf{j},\textbf{i}}$ gives rise to the matrix $\mathtens{X}_{\footnotesize\textbf{j}(n),\textbf{i}(m)}=X_{n,m},\hspace{0.2cm}n=1:N,m=1:M$, with $N=J_1J_2\dots J_p$, $M=I_1I_2\dots I_q$.

%%%%%%%%%%%%%%%%%%%%%%%
% Linear transformations
%%%%%%%%%%%%%%%%%%%%%%%
\paragraph{Linear transformations} A two-index tensor
$\mathtens{A}_{\footnotesize\hspace{0.02cm}\textbf{i},\textbf{j}}
\in{R^{I\times{J}}}$, where 
$I=I_1\times\ldots\times{I_d}$, $J=J_1\times\ldots\times{J_e}$,
defines a \textit{linear transformation }
$\mathtens{A}:\mathbb{R}^{J}\rightarrow {\mathbb{R}^{I}}$
from $\mathbb{R}^{J}$ to $\mathbb{R}^{I}$ as the product

%%%%%%%%%% eq.2.18
\begin{equation}\label{eq2.18}
\mathtens{Y}_{\footnotesize\hspace{0.02cm}\textbf{i}}=
\mathtens{A}_{\footnotesize\hspace{0.02cm}\textbf{i},\textbf{j}}
\mathtens{X}_{\footnotesize\hspace{0.02cm}\textbf{j}}
\end{equation}

that transforms a tensor 
$\mathtens{X}_{\footnotesize\hspace{0.02cm}\textbf{j}}\in{R^{J}}$ to the tensor
$\mathtens{Y}_{\footnotesize\hspace{0.02cm}\textbf{i}}\in{R^{I}}$.
Among the possible transformations a particular role is assumed by transformations from
$\mathbb{R}^{I_1 \times I_2 \times \ldots \times I_d}$ to itself. In this particular case the tensors  $\mathtens{Y}_{\footnotesize\hspace{0.02cm}\textbf{i}}$ and $\mathtens{X}_{\footnotesize\hspace{0.02cm}\textbf{j}}$ are both in $\mathbb{R}^{I_1 \times I_2 \times \ldots \times I_d}$ and  the order-$2d$ tensor $\mathtens{A}_{\footnotesize\hspace{0.02cm}\textbf{i},\textbf{j}}$ is said to be a tensor \textit{operator} (or \textit{endomorphism}). Thus, the operator $\mathtens{A}_{\textbf{i,j}}$  establishes a transformation from
$\mathtens{L}_d$ to $\mathtens{L}_d$ and (\ref{eq2.18}) can be rewritten without ambiguity as

%%%%%%%%% eq.3.1
\begin{equation}\label{eq3.1}
\mathtens{Y} = \mathtens{A} \mathtens{X}
\end{equation}

A tensor $\mathtens{V}$ in $\mathtens{L}_d$ is said to be an \textit{eigentensor} if $\mathtens{V}\neq{0}$ and if for some scalar $\lambda$ the following equation

%%%%% eq.3.2
\begin{equation}\label{eq3.2}
\mathtens{A}\mathtens{V}=\lambda\mathtens{V}
\end{equation}
is satisfied.
The scalar $\lambda$ is known as the \textit{eigenvalue} of $\mathtens{A}$ associated with the eigentensor $\mathtens{V}$,

%%%%%%%%%%%%%%%
%%%% Self-adjoint nonnegative definite tensor operators
%%%%%%%%%%%%%%%
\paragraph{Self-adjoint non-negative definite operators}
\label{sec:theory1} A \textit{self-adjoint} operator is an operator for which the following property

%%%%%%%%5 eq.3.3
\begin{equation}\label{eq3.3}
    \left\langle \mathtens{A}\mathtens{Y}, \mathtens{Z} \right\rangle =\left\langle \mathtens{Y}, \mathtens{A}\mathtens{Z} \right\rangle 
\end{equation}
%%%%%%%%%%
holds. 

As an example, an order-$2d$ tensor 
%%%%%%%%% eq.3.4
\begin{equation}\label{eq3.4}
{\mathtens{A}}_{\footnotesize\hspace{0.02cm}\textbf{i},\textbf{j}},
\hspace{0.2cm}\textbf{i}={(i_1, \ldots, i_d)},\hspace{0.2cm} \textbf{j} = {(j_1, \ldots, j_d)}
\end{equation}
%%%%%%%%%%%%%

such that

%%%%%%%%% eq.3.5
\begin{equation}\label{eq3.5}
{\mathtens{A}}_{\footnotesize\hspace{0.02cm}\textbf{i},\textbf{j}}={\mathtens{A}}_{\footnotesize\hspace{0.02cm}\textbf{j},\textbf{i}}
\end{equation}
%%%%%%%%%%%%

is a self-adjoint operator, since it results 

%%%%%%%% eq.3.6
\begin{equation}\label{eq3.6}
    \left\langle \mathtens{A}\mathtens{Y}, \mathtens{Z} \right\rangle ={\mathtens{A}}_{\footnotesize\hspace{0.02cm}\textbf{i},\textbf{j}}{\mathtens{Y}}_{\footnotesize\hspace{0.02cm}\textbf{j}}{\mathtens{Z}}_{\footnotesize\hspace{0.02cm}\textbf{i}}={\mathtens{Y}}_{\footnotesize\hspace{0.02cm}\textbf{j}}{\mathtens{A}}_{\footnotesize\hspace{0.02cm}\textbf{j},\textbf{i}}{\mathtens{Z}}_{\footnotesize\hspace{0.02cm}\textbf{i}}=\left\langle \mathtens{Y}, \mathtens{A}\mathtens{Z} \right\rangle.
\end{equation}

In the particular case the operator acts on the order-1 tensor space $R^{I_1}$, that correspoinds to $d=1$ in (\ref{eq3.4}), the self-adjoint operator (\ref{eq3.5}) reduces to a \textit{symmetric} matrix $A_{i_1,j_1}=A_{j_1,i_1}$.\\
A \textit{non-negative definite} operator is such that the following property

%%%%%%%%%%% eq.3.7
\begin{equation}\label{eq3.7}
\left\langle \mathtens{V}, 
\mathtens{A}\mathtens{V} \right\rangle\geq0
\end{equation}

holds for every tensor $\mathtens{V}$.

An important class of operators is the class of \textit{self-adjoint non-negative definite} (\textit{SA-NND}) tensor operators that are both self-adjoint and non-negative. 
For such tensors, on the basis of the properties stated before, it follows that eigenvalues of an (\textit{SA-NND}) tensor are non-negative and the eigentensors  belonging to distinct eigenvalues are orthogonal.

Indeed, combining (\ref{eq3.2}) and (\ref{eq3.3}) yields the scalar

%%%%%%% eq. 3.8
\begin{equation}\label{eq3.8}
    \lambda=\left\langle \mathtens{V}, \mathtens{A}\mathtens{V} \right\rangle/
    {\left\langle \mathtens{V}, \mathtens{V} \right\rangle}
\end{equation}
that is non-negative due to property (\ref{eq3.7}).

As for the eigentensor orthogonality, suppose that $\mathtens{A}\mathtens{V}_1=\lambda_1\mathtens{V}_1$ and $\mathtens{A}\mathtens{V}_2=\lambda_2\mathtens{V}_2$  for $\lambda_1\neq\lambda_2$, if $\mathtens{A}$ is self-adjoint , then

%%%% eq.3.9
\begin{equation}\label{eq3.9}
    \left\langle \mathtens{A}\mathtens{V}_1, \mathtens{V}_2 \right\rangle =\lambda_1\left\langle \mathtens{V}_1,\mathtens{V}_2 \right\rangle
\end{equation}
and also

%%%%%%% eq.3.10
\begin{equation}\label{eq3.10}
    \left\langle \mathtens{A}\mathtens{V}_1, \mathtens{V}_2 \right\rangle =\left\langle \mathtens{V}_1, \mathtens{A}\mathtens{V}_2\right\rangle =\lambda_2\left\langle \mathtens{V}_1,\mathtens{V}_2 \right\rangle
\end{equation}

Therefore,  $(\lambda_1-\lambda_2)\left\langle \mathtens{V}_1,\mathtens{V}_2 \right\rangle=0$, and hence $\left\langle \mathtens{V}_1,\mathtens{V}_2 \right\rangle=0$, since $\lambda_1\neq\lambda_2$.

An SA-NND operator can be easily derived from a tensor $\mathtens{A}_{\textbf{i,j}}$ as follows
%%%%%%%% eq.3.11
\begin{equation}\label{eq3.11}
\mathtens{G}_{\textbf{i,j}}=
\mathtens{A}_{\textbf{k,i}}
\mathtens{A}_{\textbf{k,j}}
\end{equation}

The operator $\mathtens{G}_{\textbf{i,j}}$ is self-adjoint, in fact we have
%%%%%%%% eq.3.12
\begin{equation}\label{eq3.12}
\mathtens{G}_{\textbf{i,j}}=
\mathtens{A}_{\textbf{k,i}}
\mathtens{A}_{\textbf{k,j}}=
\mathtens{A}_{\textbf{k,j}}
\mathtens{A}_{\textbf{k,i}}=
\mathtens{G}_{\textbf{j,i}}
\end{equation}

The eigenvalues of $\mathtens{G}_{\textbf{i,j}}$ are non-negative, since 
%%%%%%%%%%% eq. 3.13
\begin{equation}\label{eq3.13}
\left\langle \mathtens{V}, \mathtens{G}\mathtens{V} \right\rangle=
\mathtens{V}_{\textbf{i}}
\mathtens{G}_{\textbf{i,j}}
\mathtens{V}_{\textbf{j}}=
\mathtens{A}_{\textbf{k,i}}
\mathtens{V}_{\textbf{i}}
(\mathtens{A}_{\textbf{k,j}}
\mathtens{V}_{\textbf{j}})=
\mathtens{Z}_{\textbf{k}}
\mathtens{Z}_{\textbf{k}}\geq0
\end{equation}

where $\mathtens{Z}_{\textbf{k}}=
\mathtens{A}_{\textbf{k,i}}
\mathtens{V}_{\textbf{i}}$,
so that from (\ref{eq3.8}) it results $\lambda\geq0$. The operator $\mathtens{G'}_{\textbf{i,j}}=
\mathtens{A}_{\textbf{i,k}}
\mathtens{A}_{\textbf{j,k}}$ satisfies this property as well. In particular for linear indices $\textbf{i}=i$,\hspace{0.2cm}$\textbf{j}=j$,\hspace{0.2cm}$\textbf{k}=k$, the operator reduces to the symmetric matrix 

%%%%%%%%%%%%  eq.3.14
\begin{equation}\label{eq3.14}
    {G}_{i,j}={A}_{k,i}{A}_{k,j}=
    ({A}^T_{i,k}){A}_{k,j}
\end{equation}

which can be written as $G=A^TA$. Similarly, the operator 
$\mathtens{G'}_{\textbf{i,j}}$ reduces to the matrix $G^T=AA^T$. The matrix $G$ represents the well known Gram-matrix, while matrix $G^T$ its transpose.
Thus, tensor $\mathtens{G}_{\textbf{i,j}}$ is the analog in tensor space of Gram-matrix $G$. Assuming $G$ has size $(L\times L)$ and rank $r$, i.e. $r=rank(G)$, it is well known from geometry that, since $G$ is a symmetric non-negative definite matrix, it admits $L$ eigenvalues $\{\lambda_i,\hspace{0.2cm}i=1:L\}$ such that $\lambda_1\geq\lambda_2\geq\dots\geq\lambda_r>{0}
=\lambda_{r+1}=\dots =\lambda_L$. Thus $G$ can be diagonalized  as 

%%%%%%%  eq.3.15
\begin{equation}\label{eq3.15}
G_{n,m}U_{m,p}=U_{n,q}\Lambda_{q,p},\hspace{0.2cm}q,p=1:r
\end{equation}

where $U=(u_1,\dots,u_r)$ is the matrix whose columns are the eigenvectors corresponding to non-zero eigenvalues in the matrix
$\Lambda=diag(\lambda_1,\dots,\lambda_r)$. 
Using orthogonality of matrix $U$, i.e. 
$U_{m,p}U_{m,p'}=\delta_{p,p'}$ and 
$\Lambda_{q,p}=\lambda_p\delta_{q,p}$, (\ref{eq3.15}) becomes

%%%%%%%%%%% eq.3.16
\begin{equation}\label{eq3.16}
G_{n,m}=\lambda_pU_{n,p}U_{m,p}.
\end{equation}

This relationship is called  \textit{spectral decomposition} of matrix $G$ and represents a fundamental result in matrix analysis, since both principal component analysis (PCA) and singular value decomposition (SVD) are based on this decomposition. Thus it is essential to ask whether a similar relationship can be derived for the class of (\textit{SA-NND}) tensors.

% ########SECTION 3 ############
%%%%%%%%%%%%%%%% %%%%%%%%%%%%%%%%%%%%%%%%%%%%%%%%%%%%%%%%%%%%%%%%%%%%%%%%%%%%%%%%%%
% TENSOR DECOMPOSITION
%%%%%%%%%%%%%%%%%%%%%%%%%%%%%%%%%%%%%%%%%%%%%%%%%%%%%%%%%%%%%%%%%%%%%%%%%%%%%%%%%%

\section{Tensor decomposition}
\label{sec:spectral analysis}

Having proven there is a correspondence between self-adjoint tensor operators and symmetric matrices, the aim of this section is to discover that a correspondence to the spectral decomposition 
(\ref{eq3.16}) exists in tensor space.
 To take advantage of these considerations, tensor $\mathtens{A}$  to be decomposed will be treated as a linear transformation from tensor space $\mathbb{R}^{J}$ to tensor space $\mathbb{R}^{I}$, that is $\mathtens{A}:\mathbb{R}^{J}\rightarrow {\mathbb{R}^{I}}$, which for linear indices reduces to the matrix $A$. On the basis of this assumption, two cases will be considered first: i) the tensor  
$ \mathtens{A}_{\footnotesize\textbf{i,j}} \in \mathbb{R}^{I \times I}$
is real self-adjoint operator and
ii) the tensor 
$\mathtens{A}_{\footnotesize\textbf{i,j}} \in \mathbb{R}^{I \times J}$, with $I\neq J$, is a linear transformation. Using these results, then the most general case of a generic tensor will be treated at the end of the section.

%%%%%%%%%%%%%%%%
%%%% A. DECOMPOSITION  OF A SELF-ADJOINT NON-NEGATIVE TENSOR OPERATOR
%%%%%%%%%%%%%%%%

\subsection*{A. Decomposition of a self-adjoint non-negative tensor operator}
\label{sec:theory2}
The first case will be studied here deals with an order-$2d$ tensor
$\mathtens{A}_{\footnotesize\textbf{i,j}}\in \mathbb{R}^{I \times I}$ that represents a linear operator (or endomorphism) 
$\mathtens{A}:\mathbb{R}^{I}\rightarrow {\mathbb{R}^{I}}$. Thus, with reference to such a tensor, we state the following proposition.

%%%% PROPOSITION 1 %%%%%%%%%
%%%%%%%%%%%%%%%%%%%

\paragraph{Proposition1} Let 

\begin{equation}\label{eq4A1}
\mathtens{A}_{\footnotesize\textbf{i,j}}
\in\mathbb{R}^{I \times I},
\hspace{0.2cm} \textbf{i} = {(i_1, \ldots, i_d)},
\hspace{0.2cm}\textbf{j}={(j_1, \ldots, j_d)} 
\end{equation}

be a non-negative self-adjoint order-$2d$ tensor operator with ${I=I_1 \times I_2 \times \ldots \times I_d}$, there exists a set of scalars $\lambda_1\geq\lambda_2\geq\dots\geq\lambda_r>{0}$, and a set of orthogonal tensors 
$\mathtens{U}=\{\mathtens{U}_{\footnotesize\textbf{i},p},
 \hspace{0.2cm}p=1:r\}$, such that the following decomposition 

%%%%% eq.4A2
\begin{equation}\label{eq4A2}
 \mathtens{A}_{\footnotesize\textbf{i,j}}=
 \lambda_p
   {\mathtens{U}_{\footnotesize\textbf{i},p}}
   \mathtens{U}_{\footnotesize\textbf{j},p},
   \hspace{0.2cm}p=1:r
\end{equation}
holds.

%%%%%%  PROOF %%%%%%%%%%
%%%%%%%%%%%%%%%%%%%
\paragraph{Proof} First, we want to prove that the following eigenvalue equation 

%%%%% eq.4A3
\begin{equation}\label{eq4A3}  
\mathtens{A}\mathtens{U}=\lambda\mathtens{U},
\end{equation}

where $\mathtens{A}$ is a self-adjoint operator, admits solutions with  $\mathtens{U} \in \mathbb{R}^{I}$ and $\lambda>{0}$. Using index convention introduced in Section 2,  Eq. (\ref{eq4A3}) is equivalent to the $L$ equations

%%%%% eq.4A5
\begin{equation}\label{eq4A5}   
\mathtens{A}_{\footnotesize\textbf{i,j}}
 \mathtens{U}_{\footnotesize\textbf{j}}=
 \lambda\mathtens{U}_{\textbf{i}},
 \hspace{0.3cm}\textbf{i}\in{I},
 \hspace{0.2cm}L=dim(I)=I_1I_2\dots I_d
\end{equation}

which can be written as 

%%%%% eq.4A6
\begin{equation}\label{eq4A6}
(\mathtens{A}_{\footnotesize\textbf{i,j}}  -\lambda\delta_{\footnotesize\textbf{i,j}})
\mathtens{U}_{\footnotesize\textbf{j}}=0,
\end{equation}

where the term on the left 
$\mathtens{Y}_{\textbf{i}}=
(\mathtens{A}_{\footnotesize\textbf{i,j}}  -\lambda\delta_{\footnotesize\textbf{i,j}})
\mathtens{U}_{\footnotesize\textbf{j}}
$ represents an order-d tensor. The solutions of (\ref{eq4A6}) are independent on how the terms in the equation are arranged as entries in the tensor $\mathtens{Y}_{\textbf{i}}$, meaning that (\ref{eq4A6}) is invariant to a one-to-one transformation of subscript vectors.
Assuming indices $\textbf{i},\textbf{j}$ are obtained by the inverse vector-to-linear index transformation (\ref{eq2.18}) then, substituting the vector indices
$\textbf{i}(n)=T(n,:)$,
$\textbf{j}(m)=T(m,:)$, (\ref{eq4A6})  becomes

%%%%% eq.4A7
\begin{equation}\label{eq4A7}
(\mathtens{A}_{\footnotesize\textbf{i}(n),
    \footnotesize\textbf{j}(m)}
    -\lambda\delta_{\footnotesize\textbf{i}(n),\textbf{j}(m)})\mathtens{U}_{\footnotesize\textbf{j}(m)}=0
\end{equation}

By defining the $(L\times{L})$ matrices

%%%% eq.4A8
\begin{equation}\label{eq4A8}
a_{n,m}=\mathtens{A}_{\footnotesize\textbf{i}(n),\textbf{j}(m)}, 
    \hspace{0.2cm} \delta_{n,m}=\delta_{\footnotesize\textbf{i}(n),\textbf{j}(m)}
\end{equation}

and the vector

%%%% eq.4A9
\begin{equation}\label{eq4A9} 
u_m=\mathtens{U}_{\footnotesize\textbf{j}(m)},
\end{equation}

thus (\ref{eq4A7}) can be re-arranged as 

%%%%% eq.4A10
\begin{equation}\label{eq4A10}
    (a_{n,m}-\lambda\delta_{n,m})u_m=0
\end{equation}

or equivalently

%%%%% eq.4A11
\begin{equation}\label{eq4A11}
 a_{n,m}u_m=\lambda{u_n},\hspace{0.2cm}m,n=1:L
\end{equation}

Equation (\ref{eq4A11}) is equivalent to 
%%%%%%%  eq.4A12.1
\begin{equation}\label{eq4A12.1}
    Au=\lambda{u}.
\end{equation}

where the matrix $[A]_{n,m}= a_{n,m}$ is a symmetric non-negative definite ($L\times{L}$) matrix. In fact, as $\mathtens{A}_{\textbf{i,j}}$ is self-adjoint, then it results 
%%%%%%%%% eq.4A12
\begin{equation}\label{eq4A12}
 a_{n,m}=\mathtens{A}_{\footnotesize\textbf{i}(n),\textbf{j}(m)}=
    \mathtens{A}_{\footnotesize\textbf{j}(m),\textbf{i}(n)}=
     a_{m,n}
\end{equation}

that proves matrix $ a_{n,m}$ is symmetric (or self-adjoint). In addition, since 
$\left\langle \mathtens{V}, 
\mathtens{A}\mathtens{V} \right\rangle\geq0$ for every tensor $\mathtens{V}$, it results

\begin{equation}
0\leq \mathtens{V}_{\footnotesize\textbf{j}}
\mathtens{A}_{\footnotesize\textbf{i},\textbf{j}}
\mathtens{V}_{\footnotesize\textbf{j}}=
\mathtens{V}_{\footnotesize\textbf{i}(m)}
\mathtens{A}_{\footnotesize\textbf{i}(m),\textbf{j}(n)}
\mathtens{V}_{\footnotesize\textbf{j}(n)}=
v_m{A_{m,n}}v_n
\end{equation}

for every vector $[v]_{n}=v_n$, thus proving the matrix $A$ is non-negative.

 Assuming $r=rank(A)$, a set of scalars $\lambda_1\geq\lambda_2\geq\dots\geq\lambda_r>{0}$ exist such that (\ref{eq4A12.1}) holds, thus (\ref{eq4A12.1}) can be rewritten as

%%%%%%%  eq.4A12.2
\begin{equation}\label{eq4A12.2}
    A_{n,m}U_{m,p}=U_{n,q}\Lambda_{q,p}
\end{equation}

where $U=(u_1,\dots,u_r)$ and $\Lambda=diag(\lambda_1,\dots,\lambda_r)$.
Using the inverse vector-to-linear index transformation (\ref{eq2.17}) we have 

%%%%% eq.4A13
\begin{equation}\label{eq4A13}
 U_{n(\textbf{i}),q}=
 \mathtens{U}_{\footnotesize\textbf{i},q},
 \hspace{0.3cm}
 A_{n(\textbf{i}),m(\textbf{j})}=
 \mathtens{A}_{\footnotesize\textbf{i,j}}
\end{equation}

and (\ref{eq4A12.2}) becomes 
%%%%% eq.4A14
\begin{equation}\label{eq4A14}
 \mathtens{A}_{\footnotesize\textbf{i,j}}
 \mathtens{U}_{\footnotesize\textbf{j},p}=
 \mathtens{U}_{\footnotesize\textbf{i},q}
 \Lambda_{q,p},
\end{equation}

or that is the same
%%%%% eq.4B3
\begin{equation}\label{eq4B3}
   \mathtens{A}_{\footnotesize\textbf{i,j}}
   {\mathtens{U}_{\footnotesize\textbf{j},p}}
   =\lambda_p
   {\mathtens{U}_{\footnotesize\textbf{i},p}},
   \hspace{0.2cm}p=1:r
\end{equation}

This proves that (\ref{eq4A3}) admits solutions with $\lambda>0$. To decompose $\mathtens{A}_{\footnotesize\textbf{i,j}}$, the tensor product of (\ref{eq4B3}) and $\mathtens{U}_{\footnotesize\textbf{j},p'}$
yields 
%%%%% eq.4B4
\begin{equation}\label{eq4B4}
   \mathtens{A}_{\footnotesize\textbf{i,j}}
   {\mathtens{U}_{\footnotesize\textbf{j},p}}
   \mathtens{U}_{\footnotesize\textbf{j},p'}
   =\lambda_p
   {\mathtens{U}_{\footnotesize\textbf{i},p}}
   \mathtens{U}_{\footnotesize\textbf{j},p'},
   \hspace{0.2cm}p=1:r
\end{equation}

and from orthogonality ${\mathtens{U}_{\footnotesize\textbf{j},p}}
   \mathtens{U}_{\footnotesize\textbf{j},p'}=\delta_{p,p'}$, it results

%%%%% eq.4B5
\begin{equation}\label{eq4B5}
   \mathtens{A}_{\footnotesize\textbf{i,j}}
   \delta_{p,p'}
   =\lambda_l
   {\mathtens{U}_{\footnotesize\textbf{i},p}}
   \mathtens{U}_{\footnotesize\textbf{j},p'},
   \hspace{0.2cm}p=1:r
\end{equation}
that is 
%%%%% eq.4B6
\begin{equation}\label{eq4B6}
   \mathtens{A}_{\footnotesize\textbf{i,j}}
   =\lambda_p
   {\mathtens{U}_{\footnotesize\textbf{i},p}}
   \mathtens{U}_{\footnotesize\textbf{j},p},
   \hspace{0.2cm}p=1:r
\end{equation}
and this concludes the proof.

 A pseudo-code of the procedure previously described to derive a decomposition of a self-adjoint non-negative definite operator is described in Algorithm~\ref{algorithm1}.

%%%%%%%%%%%%%%%
%%%% ALGORITHM 1
%%%%%%%%%%%%%%%
\begin{algorithm}[!htb]
\large
\caption{Decomposition of self-adjoint non-negative definite tensor operator}
\label{algorithm1}
\setstretch{1.1} %1.3
\begin{algorithmic}
\STATE{INPUT: order-$2d$ self-adjoint non-negative definite operator
$\mathtens{A}_{\footnotesize\textbf{i,j}}\in \mathbb{R}^{I \times I}, L=dim(I)$}
\STATE{$\mathtens{A}_{\footnotesize\textbf{i,j}},
\hspace{0.2cm} \textbf{i} = {(i_1, \ldots, i_d)},
\hspace{0.2cm}\textbf{j}={(j_1, \ldots, j_d)}$}
\vspace{0.2cm}
\STATE{1. Derive ($L\times{d}$) matrix $T$ defining vector-to-linear transformation}
\STATE{$\textbf{i}(n)=T(n,:),\hspace{0.2cm}
\textbf{j}(m)=T(m,:),\hspace{0.2cm}m,n=1:L$}
\vspace{0.2cm}
\STATE{2. Compute symmetric non-negative definite $(L\times L)$ matrix $A$}
\STATE{$A_{n,m}=\mathtens{A}_{\footnotesize\textbf{i}(n),\textbf{j}(m)},\hspace{0.2cm}m,n=1:L$}
\vspace{0.2cm}
\STATE{3. Solve eigenvalue equation}
\STATE{$  A_{n,m}U_{m,p}=U_{n,q}\Lambda_{q,p},\hspace{0.2cm}
q,p=1:r,\hspace{0.2cm}
\Lambda=diag(\lambda_1,\dots,\lambda_r),\hspace{0.2cm} r=rank(A)$},
\vspace{0.2cm}
\STATE{4. Convert matrix $U_{n,q}$ to tensor $\mathtens{U}_{\footnotesize\textbf{i},q}$ by inverse vector-to-linear transformation}
\STATE {$U_{n(\textbf{i}),q}=
 \mathtens{U}_{\footnotesize\textbf{i},q}$}
\vspace{0.2cm}
\STATE{OUTPUT: the decomposition}
\STATE{$\mathtens{A}_{\footnotesize\textbf{i,j}}
   =\lambda_p
   {\mathtens{U}_{\footnotesize\textbf{i},p}}
   \mathtens{U}_{\footnotesize\textbf{j},p},
   \hspace{0.2cm}p=1:r$}
\end{algorithmic}
\end{algorithm}
%%%%%%%%%%%%%%%%%%%%%

%%%%%%%%%%%%%
% B. DECOMPOSITION OF A LINEAR TENSOR TRANSFORMATION
%%%%%%%%%%%%
\subsection*{B. Decomposition of a linear tensor transformation}

The second case refers to the order-$(d+e)$ tensor $\mathtens{A}_{\footnotesize\textbf{i,j}}
\in \mathbb{R}^{I \times J}$ that represents a linear transformation $\mathtens{A}:\mathbb{R}^{J}\rightarrow {\mathbb{R}^{I}}$. Thus, we state the following proposition.

%%%% PROPOSITION 2 %%%%%%%%%
%%%%%%%%%%%%%%%%%%%

\paragraph{Proposition2} Given
%%%%%%%%%% eq.4C1
\begin{equation}\label{eq4C1}
\mathtens{A}_{\footnotesize\textbf{i,j}}
\in \mathbb{R}^{I \times J},
    \hspace{0.2cm} \textbf{i} = {(i_1, \ldots, i_d)},
    \hspace{0.2cm}\textbf{j}={(j_1, \ldots, j_e)}
\end{equation}

where ${I=I_1 \times I_2 \times \ldots \times I_d}$ and ${J=J_1 \times J_2 \times \ldots \times J_e}$, there exist sets of orthogonal tensors 
$\mathtens{U}=\{\mathtens{U}_
 {\footnotesize\textbf{i},p}
 \in \mathbb{R}^{I},
 \hspace{0.2cm}p=1:r\}$,
 $\mathtens{V}=\{\mathtens{V}_
 {\footnotesize\textbf{j},p}
 \in \mathbb{R}^{J},
 \hspace{0.2cm}p=1:r\}$, 
 and a set of scalars
$\sigma_1\geq\sigma_2\geq\dots\geq\sigma_r>{0}$, such that the following decomposition

 %%%%% eq.4C2
\begin{equation}\label{eq4C2}   
\mathtens{A}_{\footnotesize\textbf{i,j}}=
\sigma_p
 \mathtens{U}_{\textbf{i},p}
 \mathtens{V}_{\footnotesize\textbf{j},p}
 \hspace{0.2cm}, p=1:r.
\end{equation}
holds.

%%%%%%%%% PROOF
\paragraph{Proof} The tensor

%%%%%% eq.4C3
\begin{equation}\label{4C3}
\mathtens{G}_{\footnotesize\textbf{l,j}}=
\mathtens{A}_{\footnotesize\textbf{i,l}}
\mathtens{A}_{\footnotesize\textbf{i,j}}
\in \mathbb{R}^{J \times J},
\hspace{0.2cm}
\textbf{l}=(l_1,\dots,l_d)
\end{equation}
is an \textit{SA-NND} operator, thus the Proposition1 applies and the eigenvalue equation 

%%%%% eq.4C4
\begin{equation}\label{eq4C4}   
\mathtens{G}_{\footnotesize\textbf{l,j}}
\mathtens{V}_{\footnotesize\textbf{j},p}
=\mathtens{V}_{\textbf{l},q}
 \Lambda_{q,p},
 \hspace{0.2cm}q,p=1:r
\end{equation}

can be derived from (\ref{eq4A2}), where 
$\Lambda_{q,p}=diag(\lambda_1,\dots,\lambda_r)$,
and 
$\mathtens{V}=\{\mathtens{V}_
 {\footnotesize\textbf{j},p},\hspace{0.2cm}p=1:r\}$
is a set of orthogonal tensors in $\mathbb{R}^{J}$. 
Since $\lambda_i>0$  we can pose $\Lambda=S^2$ with $S_{q,p}=diag(\sigma_1,\dots,\sigma_r)$, then multiplying each side of (\ref{eq4C4}) by $\mathtens{V}_
 {\footnotesize\textbf{l},q'}$ yields

 %%%%% eq.4C5
\begin{equation}\label{eq4C5} 
\mathtens{V}_
 {\footnotesize\textbf{l},q'}
\mathtens{G}_{\footnotesize\textbf{l,j}}
\mathtens{V}_{\footnotesize\textbf{j},p}
=\mathtens{V}
_{\footnotesize\textbf{l},q'}
\mathtens{V}_{\textbf{l},q}
 S^2_{q,p}
\end{equation}

and by orthogonality of $\mathtens{V}_{\textbf{l},q}$,
that is
$\mathtens{V}_{\textbf{l},q'}
\mathtens{V}_{\textbf{l},q}
=\delta_{q',q}$,
we have

%%%%% eq.4C6
\begin{equation}\label{eq4C6} 
\mathtens{V}_
 {\footnotesize\textbf{l},q}
\mathtens{A}_{\footnotesize\textbf{i,l}}
\mathtens{A}_{\footnotesize\textbf{i,j}}
\mathtens{V}_{\footnotesize\textbf{j},p}
=S^2_{q,p}
\end{equation}.

The matrix $S_{q,p}$ is diagonal full rank, thus $S^{-1}_{q,p}$ exists and (\ref{eq4C6}) can be rewritten as

%%%%% eq.4C7
\begin{equation}\label{eq4C7} 
\mathtens{A}_{\footnotesize\textbf{i,l}}
\mathtens{V}_{\footnotesize\textbf{l},q}
S^{-1}_{q,p}
(\mathtens{A}_{\footnotesize\textbf{i,j}}
\mathtens{V}_{\footnotesize\textbf{j},p}
S^{-1}_{p,q})=\delta_{p,q}
\end{equation}

which represents the orthogonality condition for the tensors

%%%%% eq.4C8
\begin{equation}\label{eq4C8}
\mathtens{U}_{\footnotesize\textbf{i},q}
=\mathtens{A}_{\footnotesize\textbf{i,j}}
\mathtens{V}_{\footnotesize\textbf{j},p}
S^{-1}_{p,q},
\hspace{0.2cm}q=1:r.
\end{equation}
Thus 
%%%%% eq.4C9
\begin{equation}\label{eq4C9}
\mathtens{U}=\{\mathtens{U}_
 {\footnotesize\textbf{i},q},\hspace{0.2cm}q=1:r\}
\end{equation}

is a set of orthogonal tensors in $\mathbb{R}^{I}$. The inner product of two generic terms $\mathtens{U}_{\footnotesize\textbf{i},s}
$ and $\mathtens{U}_{\footnotesize\textbf{i},q}
$ gives 

%%%%% eq.4C10
\begin{equation}\label{eq4C10}
\mathtens{U}_{\footnotesize\textbf{i},s}
\mathtens{U}_{\footnotesize\textbf{i},q}
=\mathtens{U}_{\footnotesize\textbf{i},s}
\mathtens{A}_{\footnotesize\textbf{i,j}}
\mathtens{V}_{\footnotesize\textbf{j},p}
S^{-1}_{p,q}
\end{equation}

and for orthogonality, that is
$\mathtens{U}_{\textbf{i},s}
\mathtens{U}_{\textbf{i},q}
=\delta_{s,q}$, it results

%%%%% eq.4C11
\begin{equation}\label{eq4C11}
S_{q,p}
=\mathtens{U}_{\footnotesize\textbf{i},q}
\mathtens{A}_{\footnotesize\textbf{i,j}}
\mathtens{V}_{\footnotesize\textbf{j},p}
\end{equation}

Using $S_{q,p}=\sigma_p\delta_{q,p}$ and the orthogonality of $\mathtens{U}_{\footnotesize\textbf{i},q}$ and $\mathtens{V}_{\footnotesize\textbf{j},p}$,  (\ref{eq4C11}) is equivalent to

%%%%% eq.4C12
\begin{equation}\label{eq4C12}
\mathtens{A}_{\footnotesize\textbf{i,j}}
=\sigma_p\mathtens{U}_{\footnotesize\textbf{i},p}
\mathtens{V}_{\footnotesize\textbf{j},p},
\hspace{0.2cm} p=1:r
\end{equation}

and this concludes the proof.

A pseudo-code of the procedure previously described to derive a decomposition of a linear tensor transformation is described in Algorithm~\ref{algorithm2}.

%%%%%%%%%%%%%%%
%%%% ALGORITHM 2
%%%%%%%%%%%%%%%
\begin{algorithm}[!htb]
\large
\caption{Decomposition of a linear tensor transformation}
\label{algorithm2}
\setstretch{1.1} %1.3
\begin{algorithmic}

\STATE{INPUT: order-$(d+e)$ tensor
\textbf{$\mathtens{A}_{\footnotesize\textbf{i,j}}
\in \mathbb{R}^{I \times J}$}}
\STATE{$\mathtens{A}_{\footnotesize\textbf{i,j}},
\hspace{0.2cm} \textbf{i} = {(i_1, \ldots, i_d)},
\hspace{0.2cm}\textbf{j}={(j_1, \ldots, j_e)}$}
\vspace{0.2cm}

\STATE{1. Derive non-negative self-adjoint operator}
\STATE{$\mathtens{G}_{\footnotesize\textbf{l,j}}=
\mathtens{A}_{\footnotesize\textbf{i,l}}
\mathtens{A}_{\footnotesize\textbf{i,j}}
\in \mathbb{R}^{J \times J}$}
\vspace{0.2cm}

\STATE{2. Solve the eigenvalue equation using Algorithm 1}
\STATE{$\mathtens{G}_{\footnotesize\textbf{l,j}}
\mathtens{V}_{\footnotesize\textbf{j},p}
=\mathtens{V}_{\textbf{l},q}
 S^2_{q,p},
 \hspace{0.2cm}q,p=1:r,
 \hspace{0.2cm}
S_{q,p}=diag(\sigma_1,\dots,\sigma_r)$}
\vspace{0.2cm}

\STATE{3. Derive the orthogonal tensors}
\STATE{$  \mathtens{U}_{\footnotesize\textbf{i},q}
=\mathtens{A}_{\footnotesize\textbf{i,j}}
\mathtens{V}_{\footnotesize\textbf{j},p}
S^{-1}_{p,q}$}
\vspace{0.2cm}

\STATE{OUTPUT: the decomposition}
\STATE{$\mathtens{A}_{\footnotesize\textbf{i,j}}
=\sigma_p\mathtens{U}_{\footnotesize\textbf{i},p}
\mathtens{V}_{\footnotesize\textbf{j},p},
\hspace{0.2cm} p=1:r$}
\end{algorithmic}
\end{algorithm}
%%%%%%%%%%%%%%%%%%%%%

%%%%%%%%%%%%%
% C. DECOMPOSITION OF A GENERIC TENSOR 
%%%%%%%%%%%%
\subsection*{C. Decomposition of a generic tensor}

The previous result can be generalized to an order-$(d+e+f)$ tensor $\mathtens{A}_{\footnotesize\textbf{i,l,k}}
\in \mathbb{R}^{I \times J\times K}$. In this case 
$\mathtens{A}_{\footnotesize\textbf{i,l,k}}$ does not unequivocally represents a transformation. With reference to this tensor we state the following proposition.

%%%% PROPOSITION 3 %%%%%%%%%
%%%%%%%%%%%%%%%%%%%

\paragraph{Proposition3} Given
%%%%%%%%%% eq.4C13
\begin{equation}\label{eq4C13}
\mathtens{A}_{\footnotesize\textbf{i,j,k}}
\in \mathbb{R}^{I \times J\times K},
    \hspace{0.2cm} \textbf{i} = {(i_1, \ldots, i_d)},
    \hspace{0.2cm}\textbf{j}={(j_1, \ldots, j_e)},
    \hspace{0.2cm}\textbf{k}={(k_1, \ldots, k_f)}
\end{equation}

where ${I=I_1 \times I_2 \times \ldots \times I_d}$, ${J=J_1 \times J_2 \times \ldots \times J_e}$ and ${K=K_1 \times K_2 \times \ldots \times K_f}$, there exist sets of orthogonal tensors 
$\mathtens{U}=\{\mathtens{U}_{\footnotesize\textbf{i},m}
\in \mathbb{R}^{I},
\hspace{0.2cm}m=1:M\}$,
 $\mathtens{Z}=\{\mathtens{Z}_{\footnotesize\textbf{j},m}
 \in \mathbb{R}^{J},
 \hspace{0.2cm}m=1:M\}$,
 $\mathtens{W}=\{\mathtens{W}_{\footnotesize\textbf{k},m}
 \in \mathbb{R}^{K},
 \hspace{0.2cm}m=1:M\}$, and a set of scalars
$\lambda_1\geq\lambda_2\geq\dots\geq\lambda_{M}>{0}$ with $M=r_1r_2$,
 such that $\mathtens{A}_{\footnotesize\textbf{i,j,k}}$ can be decomposed as
 %%%%% eq.4C14
\begin{equation}\label{eq4C14}   
\mathtens{A}_{\footnotesize\textbf{i,j,k}}=
\lambda_m
 \mathtens{U}_{\textbf{i},m}
 \mathtens{Z}_{\footnotesize\textbf{j},m}
 \mathtens{W}_{\footnotesize\textbf{k},m},
 \hspace{0.2cm}m=1:M.
\end{equation}

%%%%%%%%% PROOF
\paragraph{Proof} The product  $\mathtens{A}_{\footnotesize\textbf{i,j,k}}
\mathtens{A}_{\footnotesize\textbf{l,j,k}}$
is a self-adjoint operator on $\mathbb{R}^I$, thus the eigenvalue equation 

%%%%% eq.4C15
\begin{equation}\label{eq4C15}   
\mathtens{A}_{\footnotesize\textbf{i,j,k}}
\mathtens{A}_{\footnotesize\textbf{l,j,k}}
\hat{\mathtens{U}}_{\footnotesize\textbf{l},p}
=\hat{\mathtens{U}}_{\textbf{i},q}
 S^2_{q,p},
 \hspace{0.2cm}q,p=1:r_1
\end{equation}
admits a set of solutions 
$\mathtens{U}=\{\hat{\mathtens{U}}_{\footnotesize\textbf{i},q},\hspace{0.2cm}q=1:r_1\}$
with $S_{q,p}=diag(\sigma_1,\dots,\sigma_{r_1})$,
$\sigma_1\geq\sigma_2\geq\dots\geq\sigma_{r_1}>{0}$.
Since the matrix $S_{q,p}$ is diagonal full-rank, thus $S^{-1}_{q,p}$ exists and multiplying by $\hat{\mathtens{U}}_{\textbf{i},q}$ (\ref{eq4C15}) can be rewritten as 
%%%%% eq.4C16
\begin{equation}\label{eq4C16}   
\mathtens{A}_{\footnotesize\textbf{i,j,k}}
\hat{\mathtens{U}}_{\textbf{i},q}
S^{-1}_{q,p}
(\mathtens{A}_{\footnotesize\textbf{l,j,k}}
\hat{\mathtens{U}}_{\footnotesize\textbf{l},p}
S^{-1}_{q,p})
=\delta_{p,q},
\end{equation}
that represents the orthogonal condition for the tensor
%%%%% eq.4C17
\begin{equation}\label{eq4C17}  
\hat{\mathtens{V}}_{\footnotesize\textbf{j,k},q}=
\mathtens{A}_{\footnotesize\textbf{l,j,k}}
\hat{\mathtens{U}}_{\textbf{l},p}
S^{-1}_{p,q}
\in{\mathbb{R}^{J\times K}}.
\end{equation}
Due to orthogonality of $\hat{\mathtens{U}}_{\textbf{l},p}$ we have

%%%%% eq.4C18
\begin{equation}\label{eq4C18}  
\mathtens{A}_{\footnotesize\textbf{i,j,k}}=
\hat{\mathtens{U}}_{\textbf{i},q}
S_{q,p}
\hat{\mathtens{V}}_{\footnotesize\textbf{j,k},p}
\end{equation}

that represents the decomposition of linear transformation $\mathtens{A}_{\footnotesize\textbf{i,j,k}}$  from ${\mathbb{R}^{J\times K}}$ to ${\mathbb{R}^{I}}$ . (\ref{eq4C18}) can be further decomposed since the product $\mathtens{V}_{\footnotesize\textbf{j,k},p}
\mathtens{V}_{\footnotesize\textbf{l,k},p}$ is a self-adjoint operator on $\mathbb{R}^{J}$, thus the eigenvalue equation 

%%%%% eq.4C19
\begin{equation}\label{eq4C19}   
\hat{\mathtens{V}}_{\footnotesize\textbf{j,k},p}
\hat{\mathtens{V}}_{\footnotesize\textbf{l,k},p}
\hat{\mathtens{Z}}_{\footnotesize\textbf{l},r}
=\hat{\mathtens{Z}}_{\textbf{j},s}
 \Lambda^2_{s,r},
 \hspace{0.2cm}r,s=1:r_2
\end{equation}

admits a set of solutions $\hat{\mathtens{Z}}=\{\hat{\mathtens{Z}}_{\footnotesize\textbf{j},s},
 \hspace{0.2cm}s=1:r_2\}$ with $\Lambda_{r,s}=diag(\gamma_1,\dots,\gamma_{r_2})$,
$\gamma_1\geq\gamma_2\geq\dots\geq\gamma_{r_2}>{0}$. The matrix 
$\Lambda^{-1}_{s,r}$ exists and (\ref{eq4C19}) can be rewritten as 

%%%%% eq.4C20
\begin{equation}\label{eq4C20}   
\hat{\mathtens{V}}_{\footnotesize\textbf{j,k},p}
\hat{\mathtens{Z}}_{\textbf{j},s}
\Lambda^{-1}_{s,r}
(\hat{\mathtens{V}}_{\footnotesize\textbf{l,k},p}
\hat{\mathtens{Z}}_{\footnotesize\textbf{l},r}
\Lambda^{-1}_{r,s})
=\delta_{r,s}
\end{equation}
that represents the orthogonality condition for the tensor
%%%%% eq.4C21
\begin{equation}\label{eq4C21}  
\hat{\mathtens{W}}_{\footnotesize\textbf{k},p,r}
=\hat{\mathtens{V}}_{\footnotesize\textbf{j,k},p}
\hat{\mathtens{Z}}_{\footnotesize\textbf{j},s}
\Lambda^{-1}_{s,r}.
\end{equation}

By deriving $\hat{\mathtens{V}}_{\footnotesize\textbf{j,k},p}$ we have 
%%%%% eq.4C22
\begin{equation}\label{eq4C22} 
\hat{\mathtens{V}}_{\footnotesize\textbf{j,k},p}
=\hat{\mathtens{Z}}_{\footnotesize\textbf{j},s}
\Lambda_{s,r}
\hat{\mathtens{W}}_{\footnotesize\textbf{k},p,r}
\end{equation}

that represents the decomposition of linear transformation $\hat{\mathtens{V}}_{\footnotesize\textbf{j,k},p}$ 
from  ${\mathbb{R}^{K}}$ to ${\mathbb{R}^{J}}$. Finally combining (\ref{eq4C18}) and (\ref{eq4C22}) yields 

 %%%%% eq.4C23
\begin{equation}\label{eq4C23}   
\mathtens{A}_{\footnotesize\textbf{i,j,k}}=
S_{q,p}\Lambda_{s,r}
 \hat{\mathtens{U}}_{\textbf{i},q}
 \hat{\mathtens{Z}}_{\footnotesize\textbf{j},s}
 \hat{\mathtens{W}}_{\footnotesize\textbf{k},p,r},
\end{equation}
and more concisely 

%%%%% eq.4C24
\begin{equation}\label{eq4C24}   
\mathtens{A}_{\footnotesize\textbf{i,j,k}}=
\sigma_p\gamma_p
 \hat{\mathtens{U}}_{\textbf{i},p}
 \hat{\mathtens{Z}}_{\footnotesize\textbf{j},r}
\hat{\mathtens{W}}_{\footnotesize\textbf{k},p,r},
\end{equation}

since $S_{q,p}=\sigma_p\delta_{q,p}$ and $\Lambda_{s,r}=\gamma_r\delta_{s,r}$.
Now, let us derive the matrix $T$ that establishes the transformation between the vector index $\textbf{u}=(p,r), p=1:r_1,r=1:r_2$ and the linear index $m=1:M$ with $M=r_1{r_2}$, such that 

\begin{equation}\label{eq4C25}
    (p,r)=\textbf{u}(m)=(u_1(m),u_2(m)).
\end{equation}
Then (\ref{eq4C24}) becomes 

%%%%% eq.4C25.1
\begin{equation}\label{eq4C25.1}   
\mathtens{A}_{\footnotesize\textbf{i,j,k}}=
\sigma_{u_1(m)}\gamma_{u_2(m)}
\hat{\mathtens{U}}_{\textbf{i},u_1(m)}
 \hat{\mathtens{Z}}_{\textbf{j},u_2(m)}
 \hat{\mathtens{W}}_{\textbf{k},u_1(m),u_2(m)},
\end{equation}

By defining 
\begin{equation*}
    \lambda_m=\sigma_{u_1(m)}\gamma_{u_2(m)}
\end{equation*}
\begin{equation*}
    \mathtens{U}_{\textbf{i},m}=
    \hat{\mathtens{U}}_{\textbf{i},u_1(m)}
\end{equation*}
\begin{equation*}
    \mathtens{Z}_{\textbf{j},m}=
    \hat{\mathtens{Z}}_{\textbf{j},u_2(m)}
\end{equation*}
\begin{equation}\label{eq4C26}
    \mathtens{W}_{\textbf{k},m}=
    \hat{\mathtens{W}}_{\textbf{k},u_1(m),u_2(m)},
\end{equation}
from (\ref{eq4C25}) we have 

\begin{equation}\label{eq4C27}
\mathtens{A}_{\footnotesize\textbf{i,j,k}}=
\lambda_m
 \mathtens{U}_{\textbf{i},m}
 \mathtens{Z}_{\footnotesize\textbf{j},m}
 \mathtens{W}_{\footnotesize\textbf{k},m},
\end{equation}
and this concludes the proof.

A pseudo-code of the procedure previously described to derive a decomposition of a generic tensor $\mathtens{A}_{\footnotesize\textbf{i,l,k}}$ is described in Algorithm~\ref{algorithm3}.

%%%%%%%%%%%%%%%
%%%% ALGORITHM 3
%%%%%%%%%%%%%%%
\begin{algorithm}[!htb]
\large
\caption{Decomposition of a generic tensor}
\label{algorithm3}
\setstretch{1.1} %1.3
\begin{algorithmic}

\STATE{INPUT: order-$(d+e+f)$ tensor
$\mathtens{A}_{\footnotesize\textbf{i,j,k}}
\in \mathbb{R}^{I \times J\times K}$}
\STATE{$\mathtens{A}_{\footnotesize\textbf{i,j,k}},
\hspace{0.2cm} \textbf{i} = {(i_1, \ldots, i_d)},
\hspace{0.2cm}\textbf{j}={(j_1, \ldots, j_e)},
\hspace{0.2cm}\textbf{k}={(k_1, \ldots, k_f)}$}
\vspace{0.2cm}

\STATE{1. Solve eigenvalue equation}
\STATE{$\mathtens{A}_{\footnotesize\textbf{i,j,k}}
\mathtens{A}_{\footnotesize\textbf{l,j,k}}
\hat{\mathtens{U}}_{\footnotesize\textbf{l},p}
=\hat{\mathtens{U}}_{\textbf{i},q}
 S^2_{q,p},
 \hspace{0.2cm}q,p=1:r_1$}
\vspace{0.2cm}

\STATE{2. Derive the orthogonal tensors}
\STATE{$ \hat{\mathtens{V}}_{\footnotesize\textbf{j,k},q}=
\mathtens{A}_{\footnotesize\textbf{l,j,k}}
\hat{\mathtens{U}}_{\textbf{l},p}
S^{-1}_{p,q}
\in{\mathbb{R}^{J\times K}}$}
\vspace{0.2cm}

\STATE{3. Solve the eigenvalue equation}
\STATE{$ \hat{\mathtens{V}}_{\footnotesize\textbf{j,k},p}
\hat{\mathtens{V}}_{\footnotesize\textbf{l,k},p}
\hat{\mathtens{Z}}_{\footnotesize\textbf{l},r}
=\hat{\mathtens{Z}}_{\textbf{j},s}
 \Lambda^2_{s,r},
 \hspace{0.2cm}r,s=1:r_2$}
\vspace{0.2cm}

\STATE{4. Derive the orthogonal tensors}
\STATE{$ \hat{\mathtens{W}}_{\footnotesize\textbf{k},p,r}
=\hat{\mathtens{V}}_{\footnotesize\textbf{j,k},p}
\hat{\mathtens{Z}}_{\footnotesize\textbf{j},s}
\Lambda^{-1}_{s,r}$}
\vspace{0.2cm}

\STATE{5. Decompose the tensor $\mathtens{A}$}
\STATE{$ \mathtens{A}_{\footnotesize\textbf{i,j,k}}=
\sigma_p\gamma_p
 \hat{\mathtens{U}}_{\textbf{i},p}
 \hat{\mathtens{Z}}_{\footnotesize\textbf{j},r}
\hat{\mathtens{W}}_{\footnotesize\textbf{k},p,r}$}
\vspace{0.2cm}

\STATE{6. Apply the inverse vector-to-linear transformation to the vector index $\textbf{u}$}
\STATE{$ (p,r)=\textbf{u}(m)=(u_1(m),u_2(m))$}
\vspace{0.2cm}

\STATE{7. Define}
\STATE{$ 
    \lambda_m=\sigma_{u_1(m)}\gamma_{u_2(m)}$} 
\STATE{$\mathtens{U}_{\textbf{i},m}=
    \hat{\mathtens{U}}_{\textbf{i},u_1(m)}$}
\STATE{$\mathtens{Z}_{\textbf{j},m}=
    \hat{\mathtens{Z}}_{\textbf{j},u_2(m)}$}
\STATE{$\mathtens{W}_{\textbf{k},m}=
    \hat{\mathtens{W}}_{\textbf{k},u_1(m),u_2(m)}$}  
\vspace{0.2cm}

\STATE{OUTPUT: the decomposition}
\STATE{$\mathtens{A}_{\footnotesize\textbf{i,j,k}}=
\lambda_m
 \mathtens{U}_{\textbf{i},m}
 \mathtens{Z}_{\footnotesize\textbf{j},m}
 \mathtens{W}_{\footnotesize\textbf{k},m}$}
\end{algorithmic}
\end{algorithm}
%%%%%%%%%%%%%%%%%%%%%

% ########SECTION 4 ############
%%%%%%%%%%%%%%%% %%%%%%%%%%%%%%%%%%%%%%%%%%%%%%%%%%%%%%%%%%%%%%%%%%%%%%%%%%%%%%%%%%
% Numerical experiments
%%%%%%%%%%%%%%%%%%%%%%%%%%%%%%%%%%%%%%%%%%%%%%%%%%%%%%%%%%%%%%%%%%%%%%%%%%%%%%%%%%
\section{Numerical experiments }
\label{sec:theory3}

The aim of this section is to present some numerical experiments to validate the theoretical results previously derived for decomposition of tensor transformations.
The first experiment refers to an order-2$\textit{d}$ tensor $\mathtens{A}_{\footnotesize\textbf{i,j}}
\in \mathbb{R}^{I \times I}$ with $d=3$ and $I=[16,16,3]$, to validate the decomposition of an \textit{SA-NND}  tensor operator, proven in Proposition1. The entries of tensor $\mathtens{A}$ have been randomly generated and the behavior of eigenvalues as obtained by Algorithm 1 are shown in Fig.1. As you can see all the eigenvalues are positive and they decreases rapidly as the index $n$ increases. Tensor $\mathtens{A}$ can be reconstructed by decomposition (\ref{eq4A2}) to less than a numerical error of $2.6485\times10^{-10}$.
The second experiment deals with the decomposition of an order-$(1+2)$ tensor  $\mathtens{A}_{\footnotesize\textbf{i,j}}
\in \mathbb{R}^{I \times J}$, with $I=[64]$ and $J=[8,4]$. In this case the decomposition is obtained by applying Algorithm 2. Fig.2 shows the behavior of eigenvalues and using the decomposition (\ref{eq4C4}) the tensor $\mathtens{A}$ can be reconstructed to less than a numerical error of $2.4618\times 10^{-10}$.
Finally the third experiment refers to a tensor 
$\mathtens{A}_{\footnotesize\textbf{i,j,k}}
\in \mathbb{R}^{I \times J\times K}$ with $I=[64]$, $J=[16]$, $K=[3]$ and the decomposition described in Algorithm 3. The behavior of eigenvalues is shown in Fig.3 and the reconstruction of tensor $\mathtens{A}$ is obtained with an error of $5.2411\times 10^{-14}$.

%%%%% FIGURE - eigenvalues case 1
\begin{figure}[ht]
\includegraphics[width=0.9\linewidth,height=8cm]{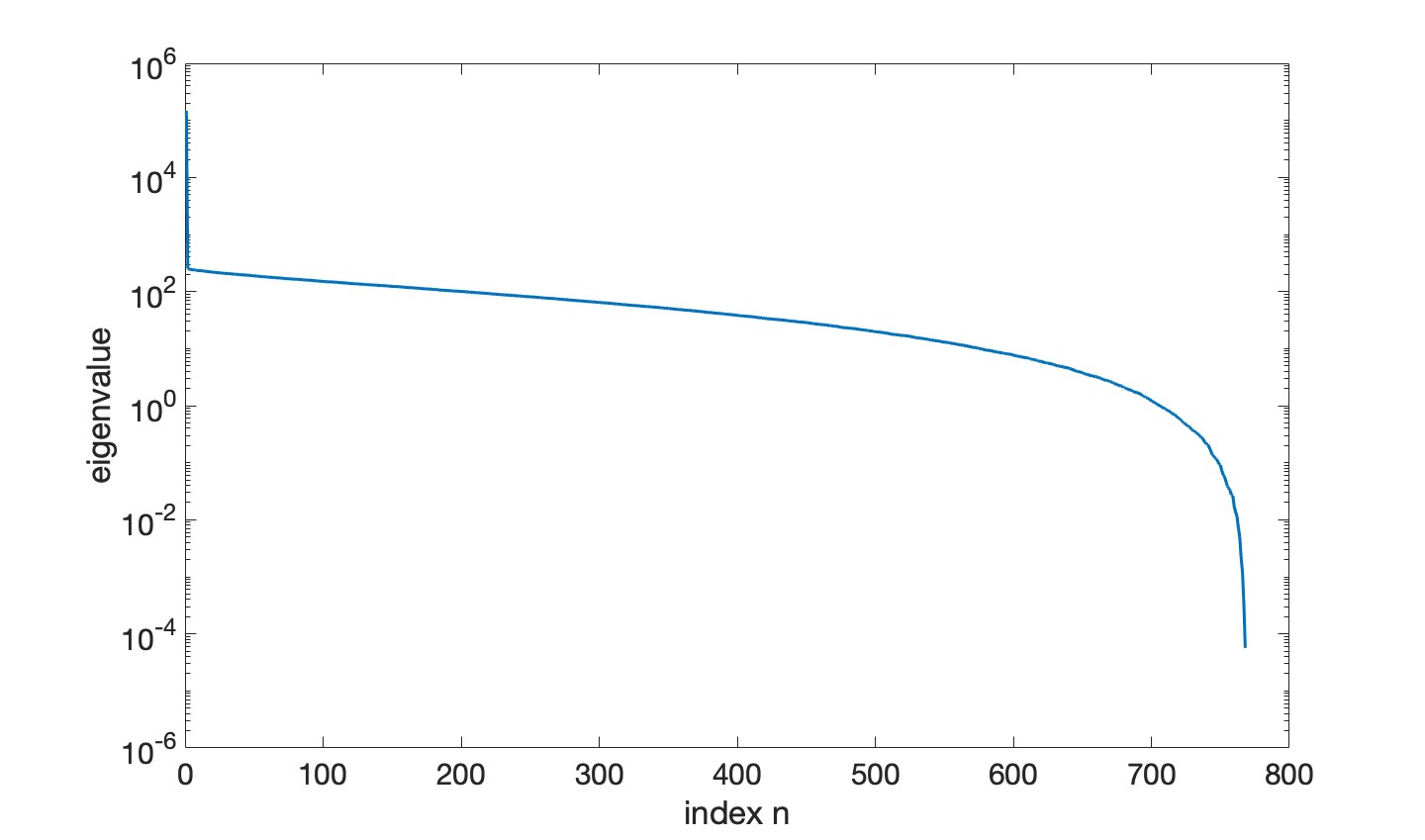}
\caption{\large{The behavior of eigenvalues as obtained from eq.(\ref{eq4A12.2})}}
\label{fig:case1eigenvalues}
\end{figure}
%%%%%%%%

%%%%% FIGURE - eigenvalues case 2
\begin{figure}[ht]
\includegraphics[width=0.9\linewidth,height=8cm]{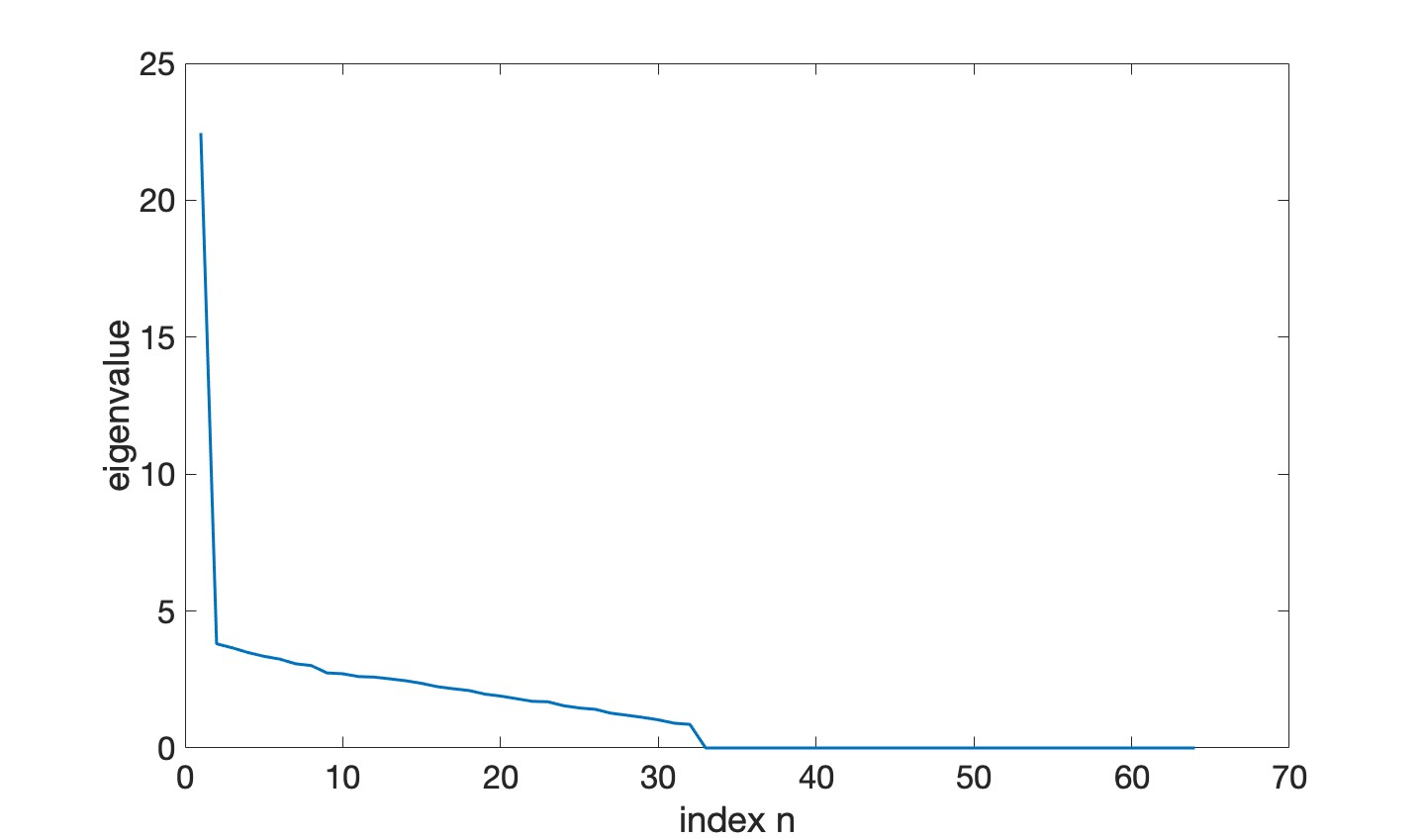}
\caption{\large{The behavior of eigenvalues as obtained from eq.(\ref{eq4C4}) }}
\label{fig:case2eigenvalues}
\end{figure}
%%%%%%%%

%%%%% FIGURE - eigenvalues case 3
\begin{figure}[ht]
\includegraphics[width=0.9\linewidth,height=8cm]{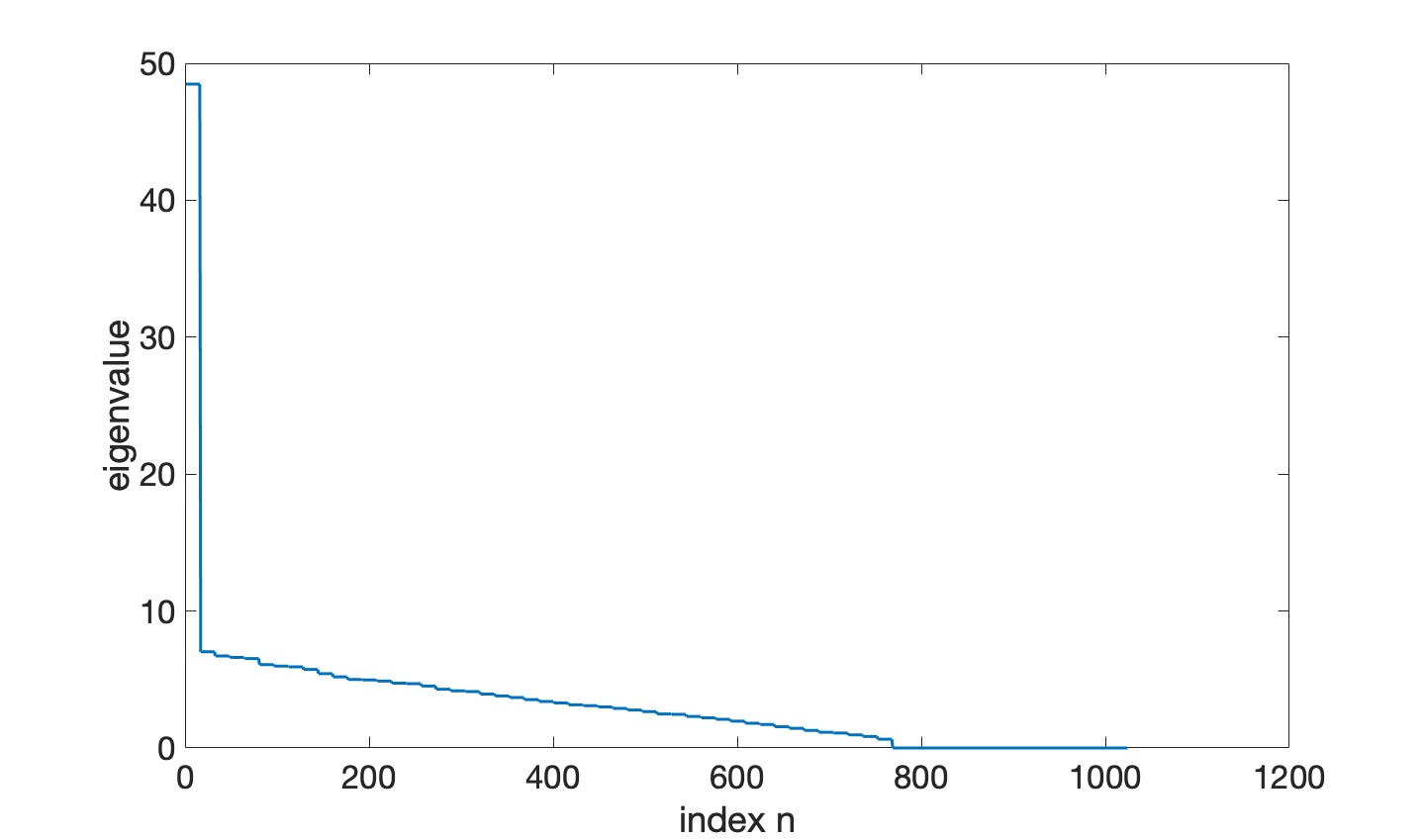}
\caption{\large{The behavior of eigenvalues as obtained from eq.(\ref{eq4C15})}}
\label{fig:case3eigenvalues}
\end{figure}
%%%%%%%%

\newpage

\section{Conclusion}
In this paper a mathematical framework for exact tensor decomposition, that is able to represent a tensor as the sum of a finite number of low-rank tensors, has been developed. The core of the proposed approach is the derivation of a decomposition for non-negative self-adjoint tensor operators. In particular it has been proven that a correspondence exists between the spectral decomposition of a symmetric matrix and the decomposition of a self-adjoint tensor operator. 

\newpage

 %%%%%%%%%%%%%%%%%%%%%%%%%%%%%%%%%%%%%%%%%%%%%%%%%%%%%%%%%%%%%%%%%%%%%%%%%%%%%%%%%%%%
% References
%%%%%%%%%%%%%%%%%%%%%%%%%%%%%%%%%%%%%%%%%%%%%%%%%%%%%%%%%%%%%%%%%%%%%%%%%%%%%%%%%%%%
%\balance
\clearpage
%\bibliographystyle{IEEEtran}
%\bibliography{IEEEabrv,references}

\begin{thebibliography}{10}
\providecommand{\url}[1]{#1}
\csname url@samestyle\endcsname
\providecommand{\newblock}{\relax}
\providecommand{\bibinfo}[2]{#2}
\providecommand{\BIBentrySTDinterwordspacing}{\spaceskip=0pt\relax}
\providecommand{\BIBentryALTinterwordstretchfactor}{4}
\providecommand{\BIBentryALTinterwordspacing}{\spaceskip=\fontdimen2\font plus
\BIBentryALTinterwordstretchfactor\fontdimen3\font minus
  \fontdimen4\font\relax}
\providecommand{\BIBforeignlanguage}[2]{{%
\expandafter\ifx\csname l@#1\endcsname\relax
\typeout{** WARNING: IEEEtran.bst: No hyphenation pattern has been}%
\typeout{** loaded for the language `#1'. Using the pattern for}%
\typeout{** the default language instead.}%
\else
\language=\csname l@#1\endcsname
\fi
#2}}
\providecommand{\BIBdecl}{\relax}
\BIBdecl

\bibitem{smilde2005multi}
A.~K. Smilde, P.~Geladi, and R.~Bro, \emph{Multi-way analysis: applications in
  the chemical sciences}.\hskip 1em plus 0.5em minus 0.4em\relax John Wiley \&
  Sons, 2005.

\bibitem{de1998matrix}
L.~De~Lathauwer and B.~De~Moor, ``From matrix to tensor: Multilinear algebra
  and signal processing,'' in \emph{Institute of mathematics and its
  applications conference series}, vol.~67.\hskip 1em plus 0.5em minus
  0.4em\relax Citeseer, 1998, pp. 1--16.

\bibitem{acar2008unsupervised}
E.~Acar and B.~Yener, ``Unsupervised multiway data analysis: A literature
  survey,'' \emph{IEEE transactions on knowledge and data engineering},
  vol.~21, no.~1, pp. 6--20, 2008.

\bibitem{tucker1966some}
L.~R. Tucker, ``Some mathematical notes on three-mode factor analysis,''
  \emph{Psychometrika}, vol.~31, no.~3, pp. 279--311, 1966.

\bibitem{carroll1970analysis}
J.~D. Carroll and J.-J. Chang, ``Analysis of individual differences in
  multidimensional scaling via an n-way generalization of “eckart-young”
  decomposition,'' \emph{Psychometrika}, vol.~35, no.~3, pp. 283--319, 1970.

\bibitem{appellof1981strategies}
C.~J. Appellof and E.~R. Davidson, ``Strategies for analyzing data from video
  fluorometric monitoring of liquid chromatographic effluents,''
  \emph{Analytical Chemistry}, vol.~53, no.~13, pp. 2053--2056, 1981.

\bibitem{sofuoglu2021multi}
S.~E. Sofuoglu and S.~Aviyente, ``Multi-branch tensor network structure for
  tensor-train discriminant analysis,'' \emph{IEEE Transactions on Image
  Processing}, vol.~30, pp. 8926--8938, 2021.

\bibitem{qin2022low}
W.~Qin, H.~Wang, F.~Zhang, J.~Wang, X.~Luo, and T.~Huang, ``Low-rank high-order
  tensor completion with applications in visual data,'' \emph{IEEE Transactions
  on Image Processing}, vol.~31, pp. 2433--2448, 2022.

\bibitem{wang2021multi}
J.-L. Wang, T.-Z. Huang, X.-L. Zhao, T.-X. Jiang, and M.~K. Ng,
  ``Multi-dimensional visual data completion via low-rank tensor representation
  under coupled transform,'' \emph{IEEE Transactions on Image Processing},
  vol.~30, pp. 3581--3596, 2021.

\bibitem{sun2022tensor}
L.~Sun, C.~He, Y.~Zheng, Z.~Wu, and B.~Jeon, ``Tensor cascaded-rank
  minimization in subspace: a unified regime for hyperspectral image low-level
  vision,'' \emph{IEEE Transactions on Image Processing}, vol.~32, pp.
  100--115, 2022.

\bibitem{long2021bayesian}
Z.~Long, C.~Zhu, J.~Liu, and Y.~Liu, ``Bayesian low rank tensor ring for image
  recovery,'' \emph{IEEE Transactions on Image Processing}, vol.~30, pp.
  3568--3580, 2021.

\bibitem{zhang2020robust}
M.~Zhang, Y.~Gao, C.~Sun, and M.~Blumenstein, ``Robust tensor decomposition for
  image representation based on generalized correntropy,'' \emph{IEEE
  Transactions on Image Processing}, vol.~30, pp. 150--162, 2020.

\bibitem{tian2022low}
X.~Tian, K.~Xie, and H.~Zhang, ``A low-rank tensor decomposition model with
  factors prior and total variation for impulsive noise removal,'' \emph{IEEE
  Transactions on Image Processing}, vol.~31, pp. 4776--4789, 2022.

\bibitem{kaur2018tensor}
D.~Kaur, G.~S. Aujla, N.~Kumar, A.~Y. Zomaya, C.~Perera, and R.~Ranjan,
  ``Tensor-based big data management scheme for dimensionality reduction
  problem in smart grid systems: Sdn perspective,'' \emph{IEEE Transactions on
  Knowledge and Data Engineering}, vol.~30, no.~10, pp. 1985--1998, 2018.

\bibitem{ji2019survey}
Y.~Ji, Q.~Wang, X.~Li, and J.~Liu, ``A survey on tensor techniques and
  applications in machine learning,'' \emph{IEEE Access}, vol.~7, pp.
  162\,950--162\,990, 2019.

\bibitem{signoretto2014learning}
M.~Signoretto, Q.~Tran~Dinh, L.~De~Lathauwer, and J.~A. Suykens, ``Learning
  with tensors: a framework based on convex optimization and spectral
  regularization,'' \emph{Machine Learning}, vol.~94, pp. 303--351, 2014.

\bibitem{hou2017fast}
M.~Hou and B.~Chaib-draa, ``Fast recursive low-rank tensor learning for
  regression.'' in \emph{IJCAI}, 2017, pp. 1851--1857.

\bibitem{lebedev2014speeding}
V.~Lebedev, Y.~Ganin, M.~Rakhuba, I.~Oseledets, and V.~Lempitsky, ``Speeding-up
  convolutional neural networks using fine-tuned cp-decomposition,''
  \emph{arXiv preprint arXiv:1412.6553}, 2014.

\bibitem{kim2015compression}
Y.-D. Kim, E.~Park, S.~Yoo, T.~Choi, L.~Yang, and D.~Shin, ``Compression of
  deep convolutional neural networks for fast and low power mobile
  applications,'' \emph{arXiv preprint arXiv:1511.06530}, 2015.

\bibitem{lin2018holistic}
S.~Lin, R.~Ji, C.~Chen, D.~Tao, and J.~Luo, ``Holistic cnn compression via
  low-rank decomposition with knowledge transfer,'' \emph{IEEE transactions on
  pattern analysis and machine intelligence}, vol.~41, no.~12, pp. 2889--2905,
  2018.

\bibitem{sidiropoulos2000parallel}
N.~D. Sidiropoulos, R.~Bro, and G.~B. Giannakis, ``Parallel factor analysis in
  sensor array processing,'' \emph{IEEE transactions on Signal Processing},
  vol.~48, no.~8, pp. 2377--2388, 2000.

\bibitem{sankaranarayanan2015tensor}
P.~Sankaranarayanan, T.~E. Schomay, K.~A. Aiello, and O.~Alter, ``Tensor gsvd
  of patient-and platform-matched tumor and normal dna copy-number profiles
  uncovers chromosome arm-wide patterns of tumor-exclusive platform-consistent
  alterations encoding for cell transformation and predicting ovarian cancer
  survival,'' \emph{PloS one}, vol.~10, no.~4, p. e0121396, 2015.

\bibitem{omberg2007tensor}
L.~Omberg, G.~H. Golub, and O.~Alter, ``A tensor higher-order singular value
  decomposition for integrative analysis of dna microarray data from different
  studies,'' \emph{Proceedings of the National Academy of Sciences}, vol. 104,
  no.~47, pp. 18\,371--18\,376, 2007.

\bibitem{zhou2013tensor}
H.~Zhou, L.~Li, and H.~Zhu, ``Tensor regression with applications in
  neuroimaging data analysis,'' \emph{Journal of the American Statistical
  Association}, vol. 108, no. 502, pp. 540--552, 2013.

\bibitem{liu2012tensor}
J.~Liu, P.~Musialski, P.~Wonka, and J.~Ye, ``Tensor completion for estimating
  missing values in visual data,'' \emph{IEEE transactions on pattern analysis
  and machine intelligence}, vol.~35, no.~1, pp. 208--220, 2012.

\bibitem{han2022rank}
Y.~Han, R.~Chen, and C.-H. Zhang, ``Rank determination in tensor factor
  model,'' \emph{Electronic Journal of Statistics}, vol.~16, no.~1, pp.
  1726--1803, 2022.

\bibitem{kolda2009tensor}
T.~G. Kolda and B.~W. Bader, ``Tensor decompositions and applications,''
  \emph{SIAM review}, vol.~51, no.~3, pp. 455--500, 2009.

\bibitem{cichocki2015tensor}
A.~Cichocki, D.~Mandic, L.~De~Lathauwer, G.~Zhou, Q.~Zhao, C.~Caiafa, and H.~A.
  Phan, ``Tensor decompositions for signal processing applications: From
  two-way to multiway component analysis,'' \emph{IEEE signal processing
  magazine}, vol.~32, no.~2, pp. 145--163, 2015.

\bibitem{sidiropoulos2017tensor}
N.~D. Sidiropoulos, L.~De~Lathauwer, X.~Fu, K.~Huang, E.~E. Papalexakis, and
  C.~Faloutsos, ``Tensor decomposition for signal processing and machine
  learning,'' \emph{IEEE Transactions on Signal Processing}, vol.~65, no.~13,
  pp. 3551--3582, 2017.

\bibitem{chen2021introduction}
H.~Chen, S.~A. Vorobyov, H.~C. So, F.~Ahmad, and F.~Porikli, ``Introduction to
  the special issue on tensor decomposition for signal processing and machine
  learning,'' \emph{IEEE Journal of Selected Topics in Signal Processing},
  vol.~15, no.~3, pp. 433--437, 2021.

\bibitem{tucker1963implications}
L.~Tucker, ``Implications of factor analysis of three way matrices
  formeasurement of change,'' \emph{Problems inmeasuring change. Madison, Wis.:
  University of WisconsinPress}, 1963.

\bibitem{harshman1970foundations}
R.~A. Harshman \emph{et~al.}, ``Foundations of the parafac procedure: Models
  and conditions for an" explanatory" multimodal factor analysis,'' 1970.

\bibitem{inoue2016generalized}
K.~Inoue, ``Generalized tensor pca and its applications to image analysis,''
  \emph{Applied Matrix and Tensor Variate Data Analysis}, pp. 51--71, 2016.

\bibitem{liu2010generalized}
J.~Liu, S.~Chen, Z.-H. Zhou, and X.~Tan, ``Generalized low-rank approximations
  of matrices revisited,'' \emph{IEEE Transactions on Neural Networks},
  vol.~21, no.~4, pp. 621--632, 2010.

\bibitem{lu2008mpca}
H.~Lu, K.~N. Plataniotis, and A.~N. Venetsanopoulos, ``Mpca: Multilinear
  principal component analysis of tensor objects,'' \emph{IEEE transactions on
  Neural Networks}, vol.~19, no.~1, pp. 18--39, 2008.

\bibitem{mavzgut2014dimensionality}
J.~Ma{\v{z}}gut, P.~Ti{\v{n}}o, M.~Bod{\'e}n, and H.~Yan, ``Dimensionality
  reduction and topographic mapping of binary tensors,'' \emph{Pattern Analysis
  and Applications}, vol.~17, pp. 497--515, 2014.

\bibitem{panagakis2009non}
Y.~Panagakis, C.~Kotropoulos, and G.~R. Arce, ``Non-negative multilinear
  principal component analysis of auditory temporal modulations for music genre
  classification,'' \emph{IEEE Transactions on Audio, Speech, and Language
  Processing}, vol.~18, no.~3, pp. 576--588, 2009.

\bibitem{sun2008incremental}
J.~Sun, D.~Tao, S.~Papadimitriou, P.~S. Yu, and C.~Faloutsos, ``Incremental
  tensor analysis: Theory and applications,'' \emph{ACM Transactions on
  Knowledge Discovery from Data (TKDD)}, vol.~2, no.~3, pp. 1--37, 2008.

\bibitem{xu2008reconstruction}
D.~Xu, S.~Yan, L.~Zhang, S.~Lin, H.-J. Zhang, and T.~S. Huang, ``Reconstruction
  and recognition of tensor-based objects with concurrent subspaces analysis,''
  \emph{IEEE Transactions on Circuits and Systems for Video Technology},
  vol.~18, no.~1, pp. 36--47, 2008.

\bibitem{yang2004two}
J.~Yang, D.~Zhang, A.~F. Frangi, and J.-y. Yang, ``Two-dimensional pca: a new
  approach to appearance-based face representation and recognition,''
  \emph{IEEE transactions on pattern analysis and machine intelligence},
  vol.~26, no.~1, pp. 131--137, 2004.

\bibitem{ye2004generalized}
J.~Ye, ``Generalized low rank approximations of matrices,'' in
  \emph{Proceedings of the twenty-first international conference on Machine
  learning}, 2004, p. 112.

\bibitem{ye2004gpca}
J.~Ye, R.~Janardan, and Q.~Li, ``Gpca: An efficient dimension reduction scheme
  for image compression and retrieval,'' in \emph{Proceedings of the tenth ACM
  SIGKDD international conference on Knowledge discovery and data mining},
  2004, pp. 354--363.

\bibitem{bro1997parafac}
R.~Bro, ``Parafac. tutorial and applications,'' \emph{Chemometrics and
  intelligent laboratory systems}, vol.~38, no.~2, pp. 149--171, 1997.

\bibitem{faber2003recent}
N.~K.~M. Faber, R.~Bro, and P.~K. Hopke, ``Recent developments in
  candecomp/parafac algorithms: a critical review,'' \emph{Chemometrics and
  Intelligent Laboratory Systems}, vol.~65, no.~1, pp. 119--137, 2003.

\bibitem{kruskal1977three}
J.~B. Kruskal, ``Three-way arrays: rank and uniqueness of trilinear
  decompositions, with application to arithmetic complexity and statistics,''
  \emph{Linear algebra and its applications}, vol.~18, no.~2, pp. 95--138,
  1977.

\bibitem{shashua2001linear}
A.~Shashua and A.~Levin, ``Linear image coding for regression and
  classification using the tensor-rank principle,'' in \emph{Proceedings of the
  2001 IEEE Computer Society Conference on Computer Vision and Pattern
  Recognition. CVPR 2001}, vol.~1.\hskip 1em plus 0.5em minus 0.4em\relax IEEE,
  2001, pp. I--I.

\bibitem{grasedyck2010hierarchical}
L.~Grasedyck, ``Hierarchical singular value decomposition of tensors,''
  \emph{SIAM journal on matrix analysis and applications}, vol.~31, no.~4, pp.
  2029--2054, 2010.

\bibitem{grasedyck2013literature}
L.~Grasedyck, D.~Kressner, and C.~Tobler, ``A literature survey of low-rank
  tensor approximation techniques,'' \emph{GAMM-Mitteilungen}, vol.~36, no.~1,
  pp. 53--78, 2013.

\bibitem{hackbusch2009new}
W.~Hackbusch and S.~K{\"u}hn, ``A new scheme for the tensor representation,''
  \emph{Journal of Fourier analysis and applications}, vol.~15, no.~5, pp.
  706--722, 2009.

\bibitem{bengua2017matrix}
J.~A. Bengua, P.~N. Ho, H.~D. Tuan, and M.~N. Do, ``Matrix product state for
  higher-order tensor compression and classification,'' \emph{IEEE Transactions
  on Signal Processing}, vol.~65, no.~15, pp. 4019--4030, 2017.

\bibitem{chaghazardi2017sample}
M.~Chaghazardi and S.~Aeron, ``Sample, computation vs storage tradeoffs for
  classification using tensor subspace models,'' \emph{arXiv preprint
  arXiv:1706.05599}, 2017.

\bibitem{haastad1989tensor}
J.~H{\aa}stad, ``Tensor rank is np-complete,'' in \emph{Automata, Languages and
  Programming: 16th International Colloquium Stresa, Italy, July 11--15, 1989
  Proceedings 16}.\hskip 1em plus 0.5em minus 0.4em\relax Springer, 1989, pp.
  451--460.

\bibitem{zare2018extension}
A.~Zare, A.~Ozdemir, M.~A. Iwen, and S.~Aviyente, ``Extension of pca to higher
  order data structures: An introduction to tensors, tensor decompositions, and
  tensor pca,'' \emph{Proceedings of the IEEE}, vol. 106, no.~8, pp.
  1341--1358, 2018.

\bibitem{cheng2016probabilistic}
L.~Cheng, Y.-C. Wu, and H.~V. Poor, ``Probabilistic tensor canonical polyadic
  decomposition with orthogonal factors,'' \emph{IEEE Transactions on Signal
  Processing}, vol.~65, no.~3, pp. 663--676, 2016.

\end{thebibliography}

%%%%%%%%%%%%%%%%%%%%%%%%%%%%%%%%%%%%%%%%%%%%%%%%%%%%%%%%%%%%%%%%%%%%%%%%%%%%%%%%%%%%
% References
%%%%%%%%%%%%%%%%%%%%%%%%%%%%%%%%%%%%%%%%
% I riferimenti vanno
% inseriti direttamente 
% in paperArxiv.tex copiando
% e incollando 
% il file output.bbl 
% (si trova in OverLeaf su
% Logs and output files, in basso a destra)
% che rappresenta la compilazione
% del file references.bib
% 
% Il file output.bbl si può 
% aprire come file di testo
% con l'applicazione TextEdit

\end{document}